 \documentclass[pmlr,twocolumn,10pt]{jmlr} 

\usepackage{booktabs}
\usepackage{siunitx}
\usepackage[utf8]{inputenc}
\usepackage[algo2e,ruled]{algorithm2e}
\usepackage{tikz}
\usepackage{array}
\usepackage{xcolor,colortbl}
\usepackage[export]{adjustbox}
\usepackage{diagbox}
\usepackage{chngcntr}
\usepackage{newfloat}
\usepackage{ragged2e}
\usepackage{nicefrac}
\usepackage{amsfonts}       
\usepackage{microtype}      
\usepackage{multirow}
\usepackage{mathtools}
\usepackage{diagbox}
\usepackage{amsfonts}
\usepackage{hyperref}
\usepackage{doi}

\graphicspath{{images/}}

\newcommand{\aref}[1]{\hyperref[#1]{Appendix~\ref*{#1}}}
\DeclareFloatingEnvironment[fileext=lop]{Supplementary Figure}
\DeclareFloatingEnvironment[fileext=lop]{Supplementary Table}

\newcommand{\eqdef}{=\vcentcolon}

\jmlrvolume{225}
\jmlryear{2023}
\jmlrsubmitted{LEAVE UNSET}
\jmlrpublished{LEAVE UNSET}
\jmlrworkshop{Machine Learning for Health (ML4H) 2023} 

 \title[M-CHMM for Healthcare Time Series]{Mixture of Coupled HMMs for Robust Modeling of Multivariate Healthcare Time Series}

\author{%
\Name{Onur Poyraz} \Email{onur.poyraz@aalto.fi}\and
\Name{Pekka Marttinen} \Email{pekka.marttinen@aalto.fi}\\
\addr Department of Computer Science, Aalto University, Finland
}

\begin{document}

\maketitle

\begin{abstract}
Analysis of multivariate healthcare time series data is inherently challenging: irregular sampling, noisy and missing values, and heterogeneous patient groups with different dynamics violating exchangeability. In addition, interpretability and quantification of uncertainty are critically important. Here, we propose a novel class of models, a mixture of coupled hidden Markov models (M-CHMM), and demonstrate how it elegantly overcomes these challenges. To make the model learning feasible, we derive two algorithms to sample the sequences of the latent variables in the CHMM: samplers based on (i) particle filtering and (ii) factorized approximation. Compared to existing inference methods, our algorithms are computationally tractable, improve mixing, and allow for likelihood estimation, which is necessary to learn the mixture model. Experiments on challenging real-world epidemiological and semi-synthetic data demonstrate the advantages of the M-CHMM: improved data fit, capacity to efficiently handle missing and noisy measurements, improved prediction accuracy, and ability to identify interpretable subsets in the data.
\end{abstract}

\section{Introduction}
\label{section:Introduction}

\begin{figure}[t]
    \centering
    \includegraphics[width=\columnwidth]{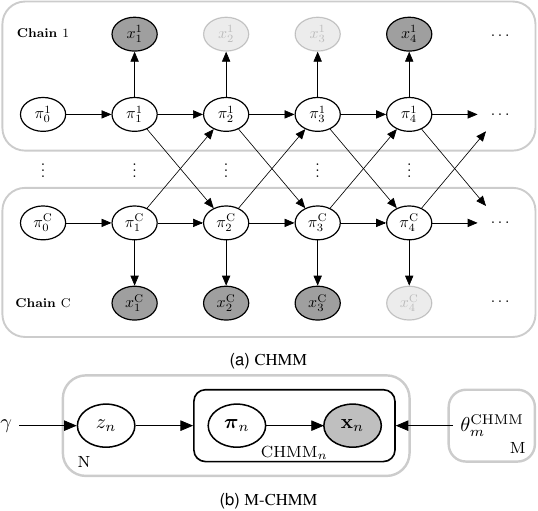}
    \caption{
    An illustration of (a) CHMM and (b) mixture of CHMMs (M-CHMM), in which each individual $n$ has a cluster label $z_n$ which specifies one of $M$ possible CHMMs that describes the dynamics of the individual.
    }
    \label{fig:chmm-mixture-graph}
\end{figure}

\begin{figure*}[t]
    \centering
    \includegraphics[width=\linewidth]{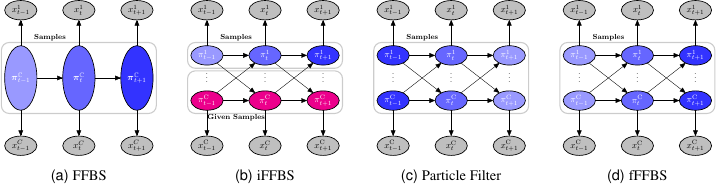}
    \caption{
    Strategies for sampling the latent variables in CHMM. The shade of {\color{blue!80}blue} shows the sampling order (darker colors are sampled first). (a) A CHMM is converted into a joint-HMM, and latent variables of chains are sampled jointly using the FFBS algorithm. (b) Latent variables of each chain are sampled conditionally on {\color{magenta}the latent variables of other chains}. (c, d) Latent variables are sampled jointly at each time step using (c) particle filtering, which results in factorized joint sampling, and (d) factorized FFBS (fFFBS), which makes the joint sampling factorizable by assumption.
    }
    \label{fig:methods}
\end{figure*}

Hidden Markov models (HMMs) \citep{rabiner1989tutorial} are widely used probabilistic frameworks for modeling longitudinal structures when some underlying hidden process happens in time. They have been widely studied and have efficient state and parameter estimation algorithms \citep{bishop2006pattern, murphy2012machine, barber2010graphical}. HMMs have a hidden state at each time point, representing the system's status at that particular time. However, many systems in the real world result from multiple interacting processes (i.e., coupled latent Markov chains), causing the state-space to grow exponentially with respect to the number of chains; therefore, learning the HMM becomes intractable \citep{touloupou2020scalable}. Coupled hidden Markov models  (CHMMs, \citet{brand1997coupled}), an extension of HMM, model interactions among multiple chains within individuals by allowing the current latent state of a chain to depend on the previous states of all the chains, as illustrated in \autoref{fig:chmm-mixture-graph}{\color{blue}a}. However, the system is still computationally intractable when the number of chains grows.

Real-world problems in epidemiology often involve longitudinal data on multiple interacting processes \citep{brand1997coupled, brand1997coupledb}. The data can be complicated by missing, irregularly collected, and noisy observations. Hence, modeling the data is non-trivial, and application of recent neural network methods for longitudinal health data \citep{alaa2022ice,kodialam21,choi2016doctor} is infeasible, especially when data are sparse and limited, as in the present work. On the other hand, probabilistic CHMMs provide two advantages for dealing with these challenges: 1) When the collected observations (e.g., medical samples or sensory information) are noisy or missing, a CHMM can model the true state as a latent chain, in which the sensitivity and specificity are available as emission parameters, and 2) the dynamic relations between the variables in the dataset are intuitively modeled by a transition matrix that describes how different latent chains affect each other. Hence, the CHMM provides a flexible, interpretable, and data-efficient alternative for modeling complex longitudinal epidemiological data. However, another challenge in real-world data is that individuals might follow different dynamics; for example, patients may respond to treatments differently, and the standard CHMM fails to capture this.

Multiple methods for inference with CHMMs have been published, which can be broadly categorized into three main branches: 1) Maximum likelihood, 2) variational approximation, and 3) Markov chain Monte Carlo (MCMC). Earlier works \citep{brand1997coupled, saul1999mixed, rezek2000learning, zhong2001new} focused primarily on the first solution; however, exact inference for CHMMs using expectation-maximization (EM) results in computational issues. \citet{wang2019variational} proposed a variational EM to alleviate this. However, our focus here is on MCMC as it is the only method that fully estimates the uncertainty of hidden states and parameters, which is important in many applications, e.g., healthcare. The main bottleneck in MCMC is the sampling of the latent chains. One straightforward solution is to convert the CHMM into a single HMM where the single latent state jointly represents the states of all the chains, and apply the standard forward filtering backward sampling (FFBS) to sample the latent variables (\autoref{fig:methods}{\color{blue}a}). However, this simple solution is infeasible with a large number of chains. Therefore, several alternative methods have been proposed. Conditional single-site \citep{dong2012graph} or block updates \citep{spencer2015super} for latent sequences provide computationally less demanding methods but produce highly correlated samples. Individual FFBS (iFFBS, \citet{touloupou2020scalable}) achieves efficiency by updating each latent chain conditionally on the other chains (\autoref{fig:methods}{\color{blue}b}); however, this conditioning increases correlation between posterior samples and does not allow for an easy way to estimate the likelihood, necessary with more complex models. 

\textit{Our first contribution} is to extend the class of CHMM to mixtures of CHMMs (M-CHMM), which resolves the challenge of distinctly behaving subgroups in healthcare time series data by learning the groups and the corresponding component models (\autoref{fig:chmm-mixture-graph}{\color{blue}b}). \textit{Our second contribution} is to investigate the suitability of two novel algorithms to sample the latent variables in CHMM: particle filtering (PF, \autoref{fig:methods}{\color{blue}c}) and factorized FFBS (fFFBS, \autoref{fig:methods}{\color{blue}d}). They (\textit{i}) are efficient and scale to a large number of states and chains, (\textit{ii}) avoid the conditioning on the latent variables of other chains, which reduces auto-correlation and improves mixing, and (\textit{iii}) allow accurate likelihood estimation by integrating out the latent variables, which is crucial in M-CHMM. We demonstrate the usefulness of our contributions by analyzing a longitudinal multivariate dataset from a clinical trial concerning the efficiency of decolonization of Methicillin-resistant \textit{Staphylococcus aureus} (MRSA) \citep{huang2019decolonization}.

\section{Background and Notation}
\label{section:Background}

\IncMargin{1.0em}
\begin{algorithm2e}[t]
\caption{Coupled Hidden Markov Model}
\label{algorithm:chmm}
\DontPrintSemicolon
\LinesNumbered
\hspace{-1em}\KwIn{Prior of $\boldsymbol{\eta}, \boldsymbol{\tau}, \boldsymbol{\epsilon}$; Initialize $\boldsymbol{\pi}$ using priors}
\KwOut{Posterior parameters}
\For {$j$ in $\mathbb{J}$}{
    \For {$c$  in $\mathbb{C}$}{
        $\eta^{c*} \sim p\left(\eta^{c} \mid \boldsymbol{\pi}^{c}\right)$ \;
        $\hphantom{\eta^{c*}}\mathllap{\tau^{c*}} \sim p\left(\tau^{c} \mid \boldsymbol{\pi} \right)$\;
        $\hphantom{\eta^{c*}}\mathllap{\epsilon^{c*}} \sim p\left(\epsilon^{c} \mid \boldsymbol{\pi}^{c}, \mathbf{x}^{c}\right)$\;
    }
    $\color{red}\boldsymbol{\pi}^{*} \sim p\left(\boldsymbol{\pi} \mid \mathbf{x}, \boldsymbol{\eta}, \boldsymbol{\tau}, \boldsymbol{\epsilon}\right)$\;
}
\end{algorithm2e}
\DecMargin{1.0em}

In this section, we provide the necessary background information about CHMMs. We first show the model formulation and a generic MCMC algorithm for the CHMM. Then, we show how to convert a CHMM into an HMM and the algorithm for the HMM. Finally, we outline the current s-o-t-a algorithm iFFBS \citep{touloupou2020scalable}. Throughout, we use $c \in \mathbb{C} = \{1,\dots, \mathrm{C}\}$ to denote a chain (shown as superscripts), where $\mathrm{C}$ is the number of chains. $k \in \mathbb{K} = \{1,\dots, \mathrm{K}\}$ and $l \in \mathbb{L} = \{1,\dots,\mathrm{L}\}$ denote sets of latent and observed states, where $\mathrm{K}, \mathrm{L}$ are the numbers of values that the latent and observed states can take for each chain (assumed equal for different chains although this could be relaxed). Time steps are represented by $t \in \mathbb{T} = \{1, \dots ,\mathrm{T} \}$, where $\mathrm{T}$ is the number of time steps. Individuals are denoted by $n \in \mathbb{N} = \{1, \ldots, \mathrm{N}\}$, where $\mathrm{N}$ denotes the number of individuals. Subscripts represent the time step(s), except in the description of M-CHMM where they represent the individual(s). Additionally, $\eta$, $\tau$, and $\epsilon$ denote initial state, transition, and emission probabilities, respectively. Joint set $\boldsymbol{\theta}=\{\eta, \tau, \epsilon\}$ denotes all the parameters of the system. Finally, $j \in \mathbb{J} = \{1, \dots, \mathrm{J}\}$ is the iterations in the MCMC, where  $\mathrm{J}$ is the number of iterations. We emphasize that except for the mixture model, we present the models and algorithms for one individual for simplicity. 

\subsection{Coupled Hidden Markov Model}
A CHMM with a latent sequence  $\boldsymbol{\pi}$, $\pi_t^c \in \{1,\dots , \mathrm{K} \}$  and observations $\mathbf{x}$, $x_t^c \in \{1,\dots , \mathrm{L}\}$ is defined as
{\small
\begin{align}
    p\left(\boldsymbol{\pi}_{1 : \mathrm{T}}^{\mathbb{C}}, \mathbf{x}_{1 : \mathrm{T}}^{\mathbb{C}}\right) =& \Biggl[\prod_{c\in \mathbb{C}} \underbrace{p\left(\pi_{1}^{c}\right)}_{\eta^c}\Biggr] \Biggl[\prod_{t=2}^{\mathrm{T}}\prod_{c\in \mathbb{C}}\underbrace{p\left(\pi_{t}^c \mid \boldsymbol{\pi}_{t-1}^{\mathbb{C}}\right)}_{\tau^c}\Biggr] \nonumber\\
    &\times\Biggl[\prod_{t=1}^{\mathrm{T}}\prod_{c\in \mathbb{C}} \underbrace{p\left(x_{t}^c \mid \pi_{t}^c\right)}_{\epsilon^c}\Biggr]. \label{eq:chmm}
  \end{align}
}\autoref{algorithm:chmm} shows a generic Gibbs sampler to learn CHMMs. The sampling of the latent sequences ({\color{red}step~7} in \autoref{algorithm:chmm}) is important because (i) it is the key to the efficiency, and (ii) marginal likelihood can be estimated as a byproduct. Therefore, it is the primary focus of this paper.

\IncMargin{1.0em}
\begin{algorithm2e}[t]
\caption{Hidden Markov Model}
\label{algorithm:hmm}
\DontPrintSemicolon
\LinesNumbered
\hspace{-1em}\KwIn{Prior of $\eta, \tau, \epsilon$; Initialize $\boldsymbol{\pi}$ using priors}
\KwOut{Posterior parameters}
\For {$j$ in $\mathbb{J}$}{
    $\eta^{*} \sim p\left(\eta \mid \boldsymbol{\pi}\right)$ \;
    $\hphantom{\eta^{*}}\mathllap{\tau^{*}} \sim p\left(\tau \mid \boldsymbol{\pi} \right)$ \;
    $\hphantom{\eta^{*}}\mathllap{\epsilon^{*}} \sim p(\epsilon \mid \boldsymbol{\pi}, \mathbf{x})$ \;
    $\color{red}\hphantom{\eta^{*}}\mathllap{\boldsymbol{\pi}^{*}} \sim p(\boldsymbol{\pi} \mid \mathbf{x}, \eta, \tau, \epsilon)$
}
\end{algorithm2e}
\DecMargin{1.0em}

\subsection{CHMM as a joint-HMM}
The CHMM is equivalent to an HMM with a latent sequence  $\boldsymbol{\pi}$, $\boldsymbol{\pi}_t^{\mathbb{C}} \in \{1,\dots , \mathrm{K}^\mathrm{C} \}$  and observations $\mathbf{x}$, $\mathbf{x}_t^{\mathbb{C}} \in \{1,\dots , \mathrm{L}^\mathrm{C} \}$, which is defined as
{\small
\begin{align}
    p\left(\boldsymbol{\pi}_{1 : \mathrm{T}}^{\mathbb{C}}, \mathbf{x}_{1 : \mathrm{T}}^{\mathbb{C}}\right) =& \underbrace{p\left(\boldsymbol{\pi}_{1}^{\mathbb{C}}\right)}_{\eta} \Biggl[\prod_{t=2}^{\mathrm{T}} \underbrace{p\left(\boldsymbol{\pi}_{t}^{\mathbb{C}} \mid \boldsymbol{\pi}_{t-1}^{\mathbb{C}}\right)}_{\tau}\Biggr]\nonumber \\
    &\times\Biggl[\prod_{t=1}^{\mathrm{T}}\underbrace{p\left(\mathbf{x}_{t}^{\mathbb{C}} \mid \boldsymbol{\pi}_{t}^{\mathbb{C}}\right)}_{\epsilon}\Biggr]. \label{eq:hmm}
\end{align}
}\autoref{algorithm:hmm} shows the Gibbs sampler for the single HMM. An efficient way of updating the latent sequences ({\color{red}step~5} on \autoref{algorithm:hmm}) is to use the \textit{forward-filtering backward-sampling (FFBS)} algorithm.

\subsection{FFBS} 
\label{subsection:Forward-Backward}
\textit{Forward-filtering} is the estimation of the current hidden state by using all observations so far, $p\left(\boldsymbol{\pi}_t^{\mathbb{C}} \mid \mathbf{x}_{1:t}^{\mathbb{C}}\right)$ \citep{barber2010graphical} given model parameters. By defining $\alpha \left(\boldsymbol{\pi}_t^{\mathbb{C}}\right) = p\left(\boldsymbol{\pi}_{t}^{\mathbb{C}} \mid \mathbf{x}_{1 : t}^{\mathbb{C}}\right)$ it yields the following recursive equation known as {\it $\alpha$-recursion}:
{\small
\begin{align}
    \textstyle \alpha\left(\boldsymbol{\pi}_{t}^{\mathbb{C}}\right) & \propto \underbrace{\textstyle p\left(\mathbf{x}_{t}^{\mathbb{C}} \mid \boldsymbol{\pi}_{t}^{\mathbb{C}}\right)}_{\text { corrector }} \underbrace{\textstyle\sum_{\boldsymbol{\pi}_{t-1}^{\mathbb{C}}} p\left(\boldsymbol{\pi}_{t}^{\mathbb{C}} \mid \boldsymbol{\pi}_{t-1}^{\mathbb{C}}\right) \alpha\left(\boldsymbol{\pi}_{t-1}^{\mathbb{C}}\right)}_{\text { predictor }}\label{eq:filtering}
\end{align}
}which is derived in \aref{section:FFBS}. The recursion starts with $\alpha \left(\boldsymbol{\pi}_1^{\mathbb{C}}\right)$ and the filtered distribution $\alpha(\cdot)$ is propagated through to the next time-step, where it acts like a `prior' \citep{barber2010graphical, barber2012bayesian}. In \autoref{eq:filtering}, the normalization constant of $\alpha \left(\boldsymbol{\pi}_t^{\mathbb{C}}\right)$ is $\mathrm{Z}_t = p\left(\mathbf{x}_t^{\mathbb{C}}\mid \mathbf{x}_{1:t-1}^{\mathbb{C}}\right)$, and the marginal likelihood is available as a byproduct of the forward-filtering:
{\small
\begin{align}
    \mathcal{L}\left(\mathbf{x}_{1:\mathrm{T}}^{\mathbb{C}}\right)\overset{\Delta}{=}p\left(\mathbf{x}_{1:\mathrm{T}}^{\mathbb{C}}\right) = \prod_{t=1}^{\mathrm{T}} \mathrm{Z}_t.
\end{align}
}

The latent variables are sampled from:
{\small
\begin{align}
    p(\boldsymbol{\pi}_{1:\mathrm{T}}^{\mathbb{C}} \mid \mathbf{x}_{1:\mathrm{T}}^{\mathbb{C}}) &= p(\boldsymbol{\pi}_{\mathrm{T}}^{\mathbb{C}} \mid  \mathbf{x}_{1:\mathrm{T}}^{\mathbb{C}}) \prod_{t=1}^{\mathrm{T}-1} p(\boldsymbol{\pi}_t^{\mathbb{C}} \mid \boldsymbol{\pi}_{t+1}^{\mathbb{C}}, \mathbf{x}_{1:t}^{\mathbb{C}}). \label{eq:backward}
\end{align}
}The posterior sampling in \autoref{eq:backward} requires `time-reversed' transitions $p(\boldsymbol{\pi}_t^{\mathbb{C}} \mid \boldsymbol{\pi}_{t+1}^{\mathbb{C}}, \mathbf{x}_{1:t}^{\mathbb{C}})$, where
{\small
\begin{align}
    p(\boldsymbol{\pi}_t^{\mathbb{C}} \mid \boldsymbol{\pi}_{t+1}^{\mathbb{C}}, \mathbf{x}_{1:t}^{\mathbb{C}}) &\propto p(\boldsymbol{\pi}_{t+1}^{\mathbb{C}} \mid \boldsymbol{\pi}_{t}^{\mathbb{C}})  \alpha (\boldsymbol{\pi}_{t}^{\mathbb{C}}), \label{eq:backward2}
\end{align}
}which are obtained using {\it $\alpha$-recursion}s from the forward filtering \citep{barber2012bayesian}. The backward sampling starts at the end of the sequence and proceeds backwards recursively using the previously sampled latent state $\boldsymbol{\pi}_{t+1}^{\mathbb{C}} = \boldsymbol{\hat{\pi}}_{t+1}^{\mathbb{C}}$:
{\small
\begin{align}
    \boldsymbol{\hat{\pi}}_{\mathrm{T}}^{\mathbb{C}} &\sim \alpha \left(\boldsymbol{\pi}_{\mathrm{T}}^{\mathbb{C}}\right), \label{eq:sampling1} \\
    \boldsymbol{\hat{\pi}}_t^{\mathbb{C}} &\sim  p\left(\boldsymbol{\pi}_{t+1}^{\mathbb{C}} = \boldsymbol{\hat{\pi}}_{t+1}^{\mathbb{C}} \mid \boldsymbol{\pi}_t^{\mathbb{C}} \right) \alpha(\boldsymbol{\pi}_t^{\mathbb{C}}). \label{eq:sampling2}
\end{align}
}This procedure is known as forward-filtering backward-sampling \citep{barber2012bayesian}, and \autoref{fig:methods}{\color{blue}a} illustrates this algorithm. The computational complexity for one time-step is of the order $\mathcal{O}\left({\left(\mathrm{K}^\mathrm{C}\right)}^2\right)$.

\subsection{Individual FFBS (iFFBS)}
Since the chains are interconnected in a CHMM, applying the FFBS separately (not jointly) on each chain will not be valid for the CHMM. It is still usable \citep{sherlock2013coupled}, but it will underestimate the interchain effects and fail if those are strong. \citet{touloupou2020scalable} reformulated the FFBS algorithm for a single chain by conditioning the updates on the states of the other chains, resulting in a proper posterior sampling. The algorithm (iFFBS) is illustrated in \autoref{fig:methods}{\color{blue}b}. By defining $\alpha\left(\pi_{t}^{c}\right) = p\big(\pi_{t}^{c} \mid \boldsymbol{\pi}_{1:t+1}^{\mathbf{\neg c}}, \mathbf{x}_{1 : t}^{c}\big)$, where $\mathbf{\neg c} = \mathbb{C}\setminus c$ refers to all the chains except $c$ itself, the modified {\it $\alpha$-recursion} is as follows:
{\small
\begin{align}
    \alpha\left(\pi_{t}^{c}\right) \propto& \underbrace{p\left(x_{t}^{c} \mid \pi_{t}^{c}\right)}_{\text {corrector}} \underbrace{\textstyle\sum_{\pi_{t-1}^{c}} p\left(\pi_{t}^{c} \mid \boldsymbol{\pi}_{t-1}^{\mathbb{C}}\right) \alpha\left(\pi_{t-1}^{c}\right)}_{\text {predictor}} \nonumber\\
    &\times \underbrace{\textstyle\prod_{\hat{c}\in \mathbb{C}\setminus c} p\left(\pi_{t+1}^{\hat{c}} \mid \boldsymbol{\pi}_{t}^{\mathbb{C}} \right)}_{\text{modifying mass}}.
\end{align}
}Here, the \textit{modifying mass} is the only additional term compared to the FFBS for a single chain. Intuitively, it corresponds to the effect of the updated chain on the other chains in the next time step (i.e., outgoing edges from the variable of interest). The marginal likelihood of the whole system is not directly available as in the FFBS algorithm \citep{touloupou2020scalable}; here we used a heuristic way to approximate it for each chain separately using forward filtering conditionally on the other chains, and take the sum. Details of this approximation is given in \aref{section:iFFBS}.

Respectively, the posterior of the latent sequences for each chain conditionally on other chains is:
{\small
\begin{align}
    p\left(\boldsymbol{\pi}_{1:\mathrm{T}}^{c} \mid \boldsymbol{\pi}_{1:\mathrm{T}}^{\mathbf{\neg c}}, \mathbf{x}_{1:\mathrm{T}}^{c}\right) &= p\left(\pi_\mathrm{T}^{c} \mid \boldsymbol{\pi}_{1:\mathrm{T}}^{\mathbf{\neg c}}, \mathbf{x}_{1:\mathrm{T}}^{c}\right) \nonumber \\
    \times&\prod_{t=1}^{\mathrm{T}-1} p\left(\pi_t^{c} \mid \pi_{t+1}^{c}, \boldsymbol{\pi}_{1:t+1}^{\mathbf{\neg c}}, \mathbf{x}_{1:t}^{c}\right),
\end{align}
}where the `time-reversed' transitions satisfy:
{\small
\begin{align}
p\left(\pi_t^{c} \mid \pi_{t+1}^{c}, \boldsymbol{\pi}_{1:t+1}^{\mathbf{\neg c}}, \mathbf{x}_{1:t}^{c}\right) \propto p\left(\pi_{t+1}^{c} \mid \pi_t^{c}, \boldsymbol{\pi}_{t}^{\mathbf{\neg c}}\right)  \alpha\left(\pi_{t}^{c}\right).
\end{align}
}Therefore, this algorithm allows using the backward sampling procedure described in \autoref{eq:sampling1} and \autoref{eq:sampling2} for each chain given the others (for details and derivations, see \aref{section:iFFBS} and \citet{touloupou2020scalable}). The computational complexity for one time-step is $\mathcal{O}\left(\mathrm{K}^2\mathrm{C}\right)$. This approach is useful yet creates correlated latent sequence samples because of the dependency on the other chains during sampling, which decreases sampling efficiency.

\section{Methods}
\label{section:Methods}

In this section, we first formally define the Mixture of CHMMs (M-CHMM) model. Then we explain how we parametrize the transition matrix in the CHMM. Finally, we propose two novel latent sequence samplers which allow us to use M-CHMM more accurately, and one of which exploits the properties of the design of the transition matrix.

\subsection{Mixture of CHMMs (M-CHMM)}

We propose the mixture of CHMMs as a flexible model for multivariate time-series data:
{\small
\begin{align}
    p\left(\mathbf{\mathbf{x}}_{n}\right) = \sum_{m=1}^{\mathrm{M}} \gamma_m \mathcal{L}_{\text{CHMM}}\left(\mathbf{x}_{n}\mid \theta_m^{\text{CHMM}}\right),
\end{align}
}where the mixing coefficients $\gamma$ satisfy $\sum_{m=1}^{\mathrm{M}} \gamma_m=1$ and $0\leq\gamma_m \leq 1$. $\mathcal{L}_{\text{CHMM}}\left(\mathbf{\mathbf{x}}_{n}\mid \theta_m^{\text{CHMM}}\right)$ denotes the marginal likelihood for the $m$th component distribution ($\boldsymbol{\pi}_{n}$ is integrated out). For inference, we augment the model with cluster labels $z_{n} \in \mathbb{M} = \{1, \dots ,\mathrm{M} \}$, yielding an equivalent formulation:
{\small
\begin{align}
    p\left(\mathbf{x}_{n}\mid z_{n}\right)&=\mathcal{L}_{\text{CHMM}}\left(\mathbf{x}_{n}\mid \theta_{z_{n}}^{\text{CHMM}}\right), \text{and}\label{eq:component_dist}\\
    z_{n}&\sim\text{Categorical}(\gamma), \quad \forall n \in \mathbb{N}.
\end{align}
}The graphical model of the M-CHMM is shown in \autoref{fig:chmm-mixture-graph}{\color{blue}b}, and \autoref{algorithm:mixture} shows a generic Gibbs sampler. Note that updating the cluster labels $z_{n}$ requires evaluating the component distributions in \autoref{eq:component_dist} for each alternative cluster.

\IncMargin{1.5em}
\begin{algorithm2e}[t]
\caption{Mixture of CHMMs}
\label{algorithm:mixture}
\DontPrintSemicolon
\LinesNumbered
\hspace{-1.5em}\KwIn{Prior of $\gamma, \boldsymbol{\eta}, \boldsymbol{\tau}, \boldsymbol{\epsilon}$; Initialize $\boldsymbol{\pi}$ using priors}
\KwOut{Posterior parameters}
\For {$j$ in $\mathbb{J}$}{
    \For {$m$  in $\mathbb{M}$}{
        \For {$c$  in $\mathbb{C}$}{
            $\eta_{m}^{c*} \sim p\left(\eta_{m}^{c} \mid \boldsymbol{\pi}^{c}, z_n=m\right)$ \;
            $\hphantom{\eta_{m}^{c*}}\mathllap{\tau_{m}^{c*}} \sim p\left(\tau_{m}^{c} \mid \boldsymbol{\pi}, z_n=m \right)$\;
            $\hphantom{\eta_{m}^{c*}}\mathllap{\epsilon_{m}^{c*}} \sim p\left(\epsilon_{m}^{c} \mid \boldsymbol{\pi}^{c}, \mathbf{x}^{c}, z_n=m\right)$
        }
    }
    $\mathbf{z}^{*} \sim p\left(\mathbf{z} \mid  \mathbf{x}, \boldsymbol{\eta}, \boldsymbol{\tau}, \boldsymbol{\epsilon}, \gamma\right)$ \;
    $\hphantom{\mathbf{z}^{*}}\mathllap{\gamma^*} \sim p\left(\gamma \mid \mathbf{z}\right)$\;
    $\color{red}\hphantom{\mathbf{z}^{*}}\mathllap{\boldsymbol{\pi}^{*}} \sim p\left(\boldsymbol{\pi} \mid  \mathbf{x}, \boldsymbol{\eta}, \boldsymbol{\tau}, \boldsymbol{\epsilon}, \mathbf{z}\right)$ \;
}
\end{algorithm2e}
\DecMargin{1.5em}

\subsection{Design of Transition Matrix in CHMM}
\label{subsection:transition}
Defining the transition matrix is a critical part of the CHMM since all the heavy computation relies on it. There is a trade-off between sparsity, which can lead to an efficient implementation \citep{sherlock2013coupled}, and the wish to carry as much information about the other chains as possible. We follow the formulation of \citet{poyraz2022modelling} and model the dependencies between the chains with parameters $\boldsymbol{\beta}$, where $\boldsymbol{\beta}^{c} \in \boldsymbol{\beta}$ denotes all parameters related to a target chain $c$. Further, each $\beta_{k}^{c\leftarrow{\hat{c}}} \in \boldsymbol{\beta}^{c}$ is a matrix of the same size as the transition matrix describing the impact the chain $\hat{c}$ has on the target chain $c$ when $\hat{c}$ is in state $k$. We assume that the rows of $\beta_{k}^{c\leftarrow{\hat{c}}}$ sum to zero, which removes redundancy and allows a one-to-one mapping between $\boldsymbol{\beta}$ and $\boldsymbol{\tau}$. We assume that there is a baseline state $k_0$ for each covariate chain $\hat{c}$, such that if the chain $\hat{c}$ is in that state it does not affect other chains, i.e., $\beta_{k_0}^{c\leftarrow{\hat{c}}} = \mathbf{0}$. The transition matrix for the chain  $c$, at time $t$, denoted by $\tau_t^{c}$, is defined as
{\small
\begin{align}
    \mu_t^{c} &= \beta_0^{c\leftarrow{c}} + \sum_{\hat{c} \in \mathbb{C} \setminus c} \sum_{k \in \mathbb{K}} \beta_k^{c\leftarrow{\hat{c}}} \mathbb{I} \left[ \pi_{t-1}^{\hat{c}} = k \right], \label{eq:additive}\\
    \tau_t^{c} &= \sigma_{\text{row}}(\mu_t^{c}). \label{eq:softmax}
\end{align}
}Hence, $\tau_t^{c}$ is obtained in \autoref{eq:softmax} by applying a row-wise softmax operator $\sigma_{\text{row}}$ to the unnormalized transition matrix $\mu_t^{c}$. Parameter $\beta_0^{c\leftarrow{c}}$ corresponds to an intercept, and it specifies the transition matrix of the target chain $c$ when all other chains are in the baseline state $k_0$, i.e., the probabilities $p(\pi_t^c\mid \pi_{t-1}^c, \pi_{t-1}^{\hat{c}}=k_0 \: \forall \hat{c}\not=c)$. Parameters $\beta_k^{c\leftarrow{\hat{c}}}$ that represent the impact of the other chains on the target chain $c$ are added to $\beta_0^{c\leftarrow{c}}$. The design of the transition matrix in \autoref{eq:additive} and \autoref{eq:softmax} yields the following factorization over chains:
{\small
\begin{align}
    p(\pi_t^c\mid \boldsymbol{\pi}_{t-1}^{\mathbb{C}})&\propto f_0(\pi_t^c,\pi_{t-1}^{c}) \prod_{\hat{c} \in \mathbb{C} \setminus c} f_{\pi_{t-1}^{\hat{c}}}(\pi_t^c, \pi_{t-1}^{c}), \label{eq:reinterpretation}
\end{align}
}where each factor $f$ uses the previous and next time step of target chain $(\pi_t^c,\pi_{t-1}^{c})$ as inputs and is a function of the latent states of other chains $\pi_{t-1}^{\hat{c}}$, except for $f_0$ which corresponds to an intercept. The derivation is in \aref{section:transition}. $\boldsymbol{\beta}$ parameters are updated with a Metropolis-Hasting step within the Gibbs sampler as proposed by \citet{sherlock2013coupled}, see details in \aref{section:MHwithinGibbs}. All methods in this paper are implemented using this trick for a fair comparison. Detailed prior distributions are given in the \aref{section:implementation}.

\subsection{Latent Sequence Samplers}
Sampling the latent sequences is a crucial part of the MCMC inference for the CHMM, as it is the bottleneck in terms of scalability. Here, we propose two novel latent sequence samplers, which are scalable and provides an unbiased estimation of the component distribution required in M-CHMM.

\subsubsection{Particle Filter for CHMM (PF)}
Here we formulate particle filter to approximate the posterior distribution of the latent sequence, $p\left(\boldsymbol{\pi}_{1:\mathrm{T}}^{\mathbb{C}}\mid \mathbf{x}_{1:\mathrm{T}}^{\mathbb{C}}\right)$, by sequentially approximating distributions, $p\left(\boldsymbol{\pi}_{1:t}^{\mathbb{C}}\mid \mathbf{x}_{1:t}^{\mathbb{C}}\right)$, for each $t$. To approximate the posterior, particle filtering places a  weighted point mass at the locations of $\mathrm{P}$ samples \citep{cemgil2014tutorial}, or \textit{particles}, $\boldsymbol{\pi}_{1:t}^{\mathbb{C},p}$, $p\in\{1,\ldots,\mathrm{P}\}$:
{\small
\begin{align}
    p\left(\boldsymbol{\pi}_{1:t}^{\mathbb{C}}\mid \mathbf{x}_{1:t}^{\mathbb{C}}\right) &\approx \hat{p}\left(\boldsymbol{\pi}_{1:t}^{\mathbb{C}}\mid \mathbf{x}_{1:t}^{\mathbb{C}}\right) \triangleq \sum_{p=1}^\mathrm{P} \frac{w_t^p}{\sum_l w_t^l}\delta_{\boldsymbol{\pi}_{1:t}^{\mathbb{C},p}}.
\end{align}
}The weighted set of particles is constructed sequentially for $t=1,\dots,\mathrm{T}$. At time $t=1$, the standard importance sampling is applied and samples are drawn from a proposal distribution, ${\boldsymbol{\pi}_1^{\mathbb{C},p}}\sim q(\boldsymbol{\pi}_1^{\mathbb{C}}) = \prod_{c\in \mathbb{C}} q(\pi_1^{c})$. For $t>1$, the optimal proposal distribution $q(\boldsymbol{\pi}_t^{\mathbb{C}}) \propto p(\boldsymbol{\pi}_t^{\mathbb{C}} \mid \boldsymbol{\pi}_{t-1}^{\mathbb{C}})p(\mathbf{x}_t^{\mathbb{C}} \mid \boldsymbol{\pi}_t^{\mathbb{C}})$ is used. Also, for $t>1$, the systematic resampling step is applied with a probability proportional to the importance weights $w_{t-1}^p$ if the effective sample size (ESS) is below a given threshold (here 0.5, see \citet{doucet2009tutorial}). The whole procedure is as follows:
{\small
\begin{alignat}{3}
&\text{resample} \quad &a_{t-1}^p &&&\sim {\textstyle\text{Cat}\left(\nicefrac{w_{t-1}^p}{\sum_k w_{t-1}^k}\right)} \\
&\text{propose}\quad  &{\boldsymbol{\pi}_t^{\mathbb{C},p}} &&&\sim {\textstyle q\left(\boldsymbol{\pi}_t^{\mathbb{C}} \mid {\boldsymbol{\pi}_{t-1}^{\mathbb{C},a_{t-1}^p}}\right)} \\
&\text{append}\quad  &{\boldsymbol{\pi}_{1:t}^{\mathbb{C},p}} &&&= {\textstyle\left({\boldsymbol{\pi}_{1:t-1}^{\mathbb{C},a_{t-1}^p}}, {\boldsymbol{\pi}_t^{\mathbb{C},p}} \right)} \\
&\text{reweight}\quad  &w_t^p &&&= {\textstyle\frac{p\left(\boldsymbol{\pi}_t^{\mathbb{C}} \mid {\boldsymbol{\pi}_{t-1}^{\mathbb{C},a_{t-1}^p}}\right) p\left(x_t^{\mathbb{C}} \mid {\boldsymbol{\pi}_t^{\mathbb{C},p}}\right)}{q\left(\boldsymbol{\pi}_t^{\mathbb{C}} \mid {\boldsymbol{\pi}_{t-1}^{\mathbb{C},a_{t-1}^p}}\right)}}
\end{alignat}
}The final particles (samples) ${\boldsymbol{\pi}_{1:\mathrm{T}}^{\mathbb{C},p}}$ and weights $w_{\mathrm{T}}^p$ define the particle filter approximation to the posterior. An unbiased estimate of the marginal likelihood can be calculated as a side-product \citep{naesseth2018variational};
{\small
\begin{align}
    \mathcal{L}\left(\mathbf{x}_{1:\mathrm{T}}^{\mathbb{C}}\right)\overset{\Delta}{=}\hat{p}\left(\mathbf{x}_{1:\mathrm{T}}^{\mathbb{C}}\right) = \prod_{t=1}^{\mathrm{T}} \frac{1}{\mathrm{P}} \sum_{p=1}^{\mathrm{P}} w_t^p
\end{align}
}The algorithm is illustrated in \autoref{fig:methods}{\color{blue}c} and the computational complexity for one time-step is $\mathcal{O}\left(\mathrm{K}\mathrm{C}\mathrm{P}\right)$.

\subsubsection{Factorized FFBS (fFFBS)}

The complexity of the FFBS for the single HMM comes from the fact that $\alpha\left(\boldsymbol{\pi}_{t}^{\mathbb{C}}\right)$ in \autoref{eq:filtering} depends on all the latent variables, requiring a summation over all combinations. To overcome this, we assume the following factorization: 
{\small
\begin{align}
p\left(\boldsymbol{\pi}_t^{\mathbb{C}}\mid \mathbf{x}_{1:t}^{\mathbb{C}}\right) \approx \prod_{c\in \mathbb{C}} p\left(\pi_t^{c}\mid x_t^c, \mathbf{x}_{1:t-1}^{\mathbb{C}}\right), \label{eq:factorization}
\end{align}
}for all $t$. \autoref{eq:factorization} does not hold exactly because previous $\boldsymbol{\pi}_{t-1}^\mathbb{C}$ make $\pi_t^c$ dependent even conditionally on observations $\mathbf{x}_{1:t-1}^{\mathbb{C}}$ from all chains. However, it is accurate when within-chain dependencies outweigh between-chain dependencies, or when the emission probabilites reflect the underlying state accurately in which case the information in $\boldsymbol{\pi}_{t-1}^\mathbb{C}$ can be extracted from observations $\mathbf{x}_{1:t-1}^{\mathbb{C}}$. If we define $\alpha\left(\pi_{t}^c\right) = p\left(\pi_t^{c} \mid x_t^c, \mathbf{x}_{1:t-1}^{\mathbb{C}}\right)$, then $\alpha\left(\boldsymbol{\pi}_{t}^{\mathbb{C}}\right) = \prod_{c\in \mathbb{C}} \alpha\left(\pi_{t}^c\right)$. The approximated {\it $\alpha$-recursion} for each chain is given by:
{\small
\begin{align}
\alpha(\pi_{t}^c) \propto& \underbrace{p\left(x_{t}^{c} \mid \pi_{t}^{c}\right)}_{\epsilon^c} \sum_{\pi_{t-1}^{c}} \alpha(\pi_{t-1}^c) \nonumber\\
&\times\underbrace{\sum_{\pi_{t-1}^{\mathbb{C}\setminus c}} p\left(\pi_{t}^{c} \mid \boldsymbol{\pi}_{t-1}^{\mathbb{C}}\right) \prod_{\hat{c} \in \mathbb{C }\setminus c} \alpha(\pi_{t-1}^{\hat{c}})}_{\tau_t^c}, \label{eq:chain-forward}
\end{align}
}as derived in \aref{section:fFFBS}. Here, by using \autoref{eq:factorization}, the summation over all latent state combinations in \autoref{eq:filtering} is re-arranged, such that for each chain there is a summation over the latent states of the other chains inside a dynamic transition matrix $\tau_t^c$. So, $\tau_t^c$ requires the summation over $\mathrm{C}-1$ latent states of the other chains, which equals to $\mathrm{K}^{\mathrm{C}-1}$ summations. However, we can express any $\tau_t^c$ as following by exploiting the proportionality in \autoref{eq:reinterpretation}:
{\small
\begin{align}
    \tau_t^c = f_0(\pi_t^c,\pi_{t-1}^{c}) \prod_{\hat{c} \in \mathbb{C} \setminus c}  \sum_{\boldsymbol{\pi}_{t-1}^{\hat{c}}} f_{\pi_{t-1}^{\hat{c}}}(\pi_t^c, \pi_{t-1}^{c}) \alpha(\pi_{t-1}^{\hat{c}}), \label{eq:dynamic-transition}
\end{align}
}which is derived in \aref{section:fFFBS}. By plugging \autoref{eq:dynamic-transition} into \autoref{eq:chain-forward}, the big summation over all latent state combinations in \autoref{eq:filtering} is separated into independent summations for each chain. Therefore,we only need $\mathrm{KC}$ summations instead of $\mathrm{K}^{\mathrm{C}}$, which makes fFFBS scalable to a high number of states and chains. In \autoref{eq:chain-forward}, the normalization constant of $\alpha(\pi_{t}^c)$ is $\mathrm{Z}_t^c = p\left(\mathbf{x}_t^c \mid \mathbf{x}_{1:t-1}^{\mathbb{C}}\right)$, and the marginal likelihood is a byproduct of the forward-filtering under the factorization assumption:
{\small
\begin{align}
    \mathcal{L}\left(\mathbf{x}_{1:\mathrm{T}}^{\mathbb{C}}\right)\overset{\Delta}{=}p\left(\mathbf{x}_{1:\mathrm{T}}^{\mathbb{C}}\right) = \prod_{t=1}^{\mathrm{T}} \prod_{c=1}^{\mathrm{C}} \mathrm{Z}_t^c.
\end{align}
}

The latent variables are sampled from:
{\small
\begin{align}
    p(\boldsymbol{\pi}_{1:\mathrm{T}}^{\mathbb{C}} \mid \mathbf{x}_{1:\mathrm{T}}^{\mathbb{C}}) &= p(\boldsymbol{\pi}_{\mathrm{T}}^{\mathbb{C}} \mid  \mathbf{x}_{1:\mathrm{T}}^{\mathbb{C}}) \prod_{t=1}^{\mathrm{T}-1} p(\boldsymbol{\pi}_t^{\mathbb{C}} \mid \boldsymbol{\pi}_{t+1}^{\mathbb{C}}, \mathbf{x}_{1:t}^{\mathbb{C}}),
\end{align}
}where the `time-reversed' transitions satisfy:
{\small
\begin{align}
    &p(\boldsymbol{\pi}_t^{\mathbb{C}} \mid \boldsymbol{\pi}_{t+1}^{\mathbb{C}}, \mathbf{x}_{1:t}^{\mathbb{C}}) \nonumber \\
    &\qquad \propto \prod_{c\in \mathbb{C}} \alpha\left(\pi_{t}^{c}\right)  f_0(\pi_{t+1}^c,\pi_{t}^{c}) \prod_{\hat{c} \in \mathbb{C} \setminus c} f_{\pi_{t}^{c}}(\pi_{t+1}^{\hat{c}}, \pi_{t}^{\hat{c}}), \label{eq:chmm-backward}
\end{align}
}which is derived in \aref{section:fFFBS}. Similar to the FFBS, the backward sampling starts at the end $\hat{\pi}_{\mathrm{T}}^c \sim \alpha (\pi_{\mathrm{T}}^c), \forall c \in \mathbb{C}$, and proceeds recursively backwards using \autoref{eq:chmm-backward} with previously sampled $\hat{\boldsymbol{\pi}}_{t+1}^{\mathbb{C}}$, as illustrated in \autoref{fig:methods}{\color{blue}dMix}. The computational complexity for one time-step is $\mathcal{O}\left(\mathrm{K}^3\mathrm{C}\right)$.

\section{Experiments and Results}
\label{section:Results}

In this section, we first introduce datasets used in experiments and our experimental setup. Then, we show that our proposed sampling algorithms for CHMM outperform existing alternatives in terms of stability, computational efficiency, and ability to detect known clusters using the M-CHMM. Finally, we show how using the M-CHMM instead of the standard CHMM improves model fit and prediction accuracy with complex real-world data.

\subsection{Datasets and Experimental Setup}

We used a randomized controlled trial (RCT) dataset from the CLEAR (Changing Lives by Eradicating Antibiotic Resistance) \citep{huang2019decolonization}. The trial subjects were randomized into two groups (\textit{control}: $n=1063$ and \textit{treatment}: $n=1058$) to test the efficiency of a treatment protocol to clear the bacterium which colonizes the patient in different body parts (nares, throat, and skin, i.e., chains). Measurements indicating the presence or absence of a bacterium in the body parts were collected from the subjects at hospital discharge ($t_0$) and on three follow-up visits 1, 3, and 6 months after the discharge ($t_1, t_3$, and $t_6$). Measurements were noisy, and some were missing. Overall, $19671$ measurements ($10039$ and $9632$ from the control and treatment) were received. We treated skipped months ($t_2$, $t_4$, and $t_5$) as additional missing observations. The dimensions of latent and observed variables were $K=L=2$, where the latent chains corresponded to whether the bacterium was truly present and observations were the noisy measurements. Hence, emission parameters give us the sensitivity and specificity of the measurements.

We created two sets of semi-synthetic datasets using the CLEAR data. For the first set (SS-1), we trained a single HMM on each trial group using the FFBS algorithm and simulated synthetic datasets with different numbers of time points (5, 7, 10, 20, 50, and 100) using the trained models. The SS-1 is the most realistic regeneration of the original CLEAR dataset. The second set (SS-2) was simulated from two CHMMs trained on control and treatment groups, respectively, using the iFFBS algorithm (not to create any advantage to our proposed models), setting the sequence length to 7 to mimic the original data. Since this dataset is created using the CHMM formulation, we can check whether CHMM training with different samplers can learn the true underlying parameters ($\boldsymbol{\beta}$-parameters described in the \autoref{section:Methods}). The posterior predictive checkings for the data-generator models for SS-1 and SS-2 are given in the \aref{section:additional}. The semi-synthetic datasets always had 1000 individuals in each group. The single HMM formulation of the CHMM and the state-of-the-art iFFBS were used as baseline samplers.

\begin{figure}[!t]
    \centering
    \includegraphics[width=\columnwidth]{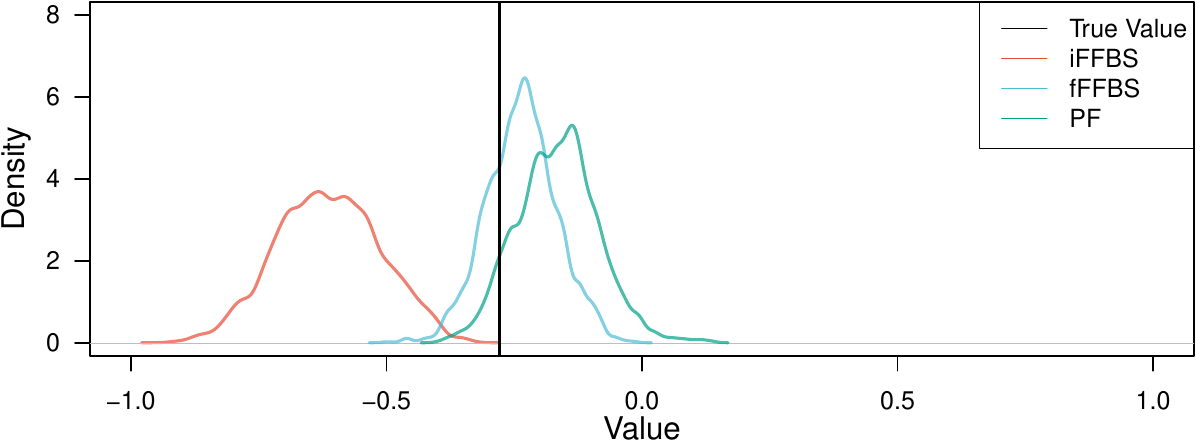}
    \caption{Estimated posterior distributions for one of the $\boldsymbol{\beta}$ parameters with different latent sequence samplers.}
    \label{fig:convergence}
\end{figure}

\subsection{Comparison of Latent Sequence Samplers}

\paragraph{Stability.}
We conducted this analysis on the SS-2 dataset. First, we compared the error in forward filtering probabilities for different algorithms, considering the single HMM as the ground truth (\aref{section:additional}). Here, we run FFBS and fFFBS just once since they can calculate forward probabilities deterministically. However, PF and iFFBS are based on samples, so they were averaged over 1000 iterations. iFFBS results in fluctuating probabilities, while the fFFBS algorithm has a slight bias due to the factorization assumption. On the other hand, PF converges to the true forward filtering probabilities. Next, we compared the estimated posterior distributions for the transition parameters ($\boldsymbol{\beta}$-parameters) on SS-2 data; one of the parameters is shown in \autoref{fig:convergence} and the rest in \aref{section:additional}. We see that fFFBS and PF are more accurate than iFFBS and that some posteriors are shrunk towards $0$ because of the prior, as expected. We note that the FFBS algorithm is excluded since it is inapplicable for this test.

\begin{figure}[b]
    \centering
    \includegraphics[width=\columnwidth]{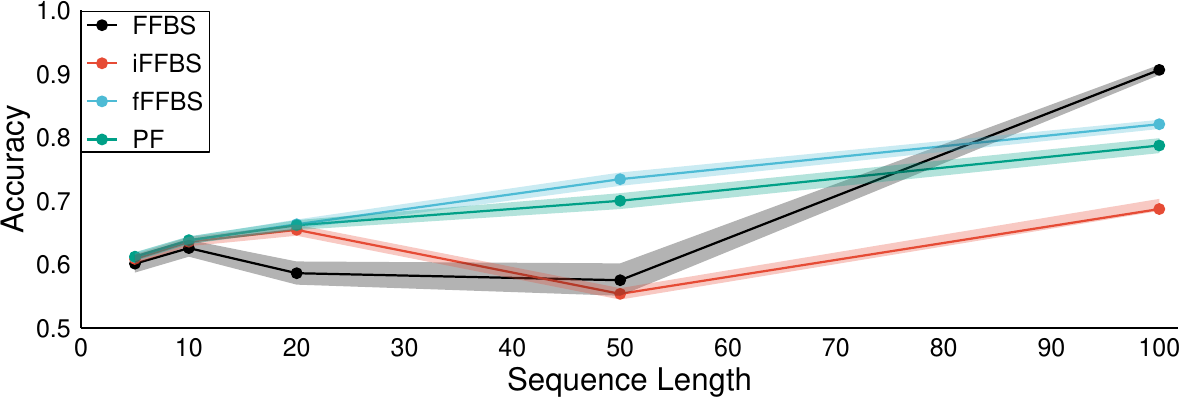}
    \caption{Clustering accuracy with different sequence lengths. Solid lines represents the mean and shaded areas the 95\% confidence interval.}
    \label{fig:accuracy-by-length}
\end{figure}

\paragraph{Computational Efficiency.}
We also compared all algorithms in terms of runtime (\autoref{table:speed}) and effective sample size (in \aref{section:additional}). This analysis shows that the single HMM is intractable when the number of chains or latent state dimension increases. The fFFBS and iFFBS algorithms scale approximately similarly when the dimensions increase. On the other hand, PF is much more efficient if the number of particles stays the same. However, a larger number of particles may be helpful with a higher dimension.

\begin{table}[!t]
        \centering
        \caption{Time comparison with different latent state and chain numbers. For each setup, sequence length T=20, a number of iterations J=100, and particle number P=10 are selected.}
        \begin{tabular}{l c c c c}
        \toprule
        \textbf{Setup} & \textbf{FFBS} & \textbf{iFFBS} & \textbf{fFFBS} & \textbf{\hspace{0.2cm}PF\hspace{0.2cm}} \\
        \cmidrule(lr){1-1}\cmidrule(lr){2-5}
        K=2, C=4 & 1.2s & 5.7s & 8.6s & 8.6s \\
        K=4, C=4 & 90s & 13s & 25s & 10s \\
        K=2, C=8 & 92s & 20s & 42s & 21s \\
        K=4, C=8 & NA & 52s & 132s & 25s \\
        K=8, C=4 & NA & 40s & 120s & 13s \\
        K=2, C=16 & NA & 109s & 239s & 63s \\
        K=8, C=8 & NA & 200s & 650s & 35s \\
        K=4, C=16 & NA & 337s & 775s & 83s \\
        K=8, C=16 & NA & 1313s & 3826s & 154s \\
        \bottomrule
        \end{tabular}
        \label{table:speed}
\end{table}

\paragraph{Ability to detect known clusters with M-CHMM.}
Here, we compared the inference algorithms with the M-CHMM in an unsupervised task by measuring how well the model learns the two clusters in the semi-synthetic SS-1 data corresponding to the control and treatment groups. The results in \autoref{fig:accuracy-by-length} show that increasing the length of the observed sequence enhances accuracy in general, and that our proposed samplers (fFFBS and PF) achieve better and more consistent results than the alternatives. This is expected because (\textit{i}) compared to the iFFBS, they can accurately estimate the likelihoods of the component models required when updating the cluster labels, and (\textit{ii}) compared to the FFBS, they benefitted from the structure within the latent sequence, which made the inference more efficient and stable especially when the sequence length is insufficient. 

\begin{table}[!t]
\centering
\caption{Comparison of models with negative log-likelihood (NLL) in the test set for control and treatment groups using 5-fold cross-validation (CV). A lower score is better.}
\begin{tabular}{l c c}
\toprule
\textbf{Model} & \textbf{Control} & \textbf{Treatment}\\
\cmidrule(lr){1-1} \cmidrule(lr){2-2} \cmidrule(lr){3-3}
\textbf{CHMM} & $999\pm44$ & $812\pm42$ \\
\textbf{M-CHMM ($\mathrm{M}=2$)} & $855\pm50$ & $663\pm31$ \\
\textbf{M-CHMM ($\mathrm{M}=3$)} & $808\pm48$ & $\mathbf{585\pm32}$ \\
\textbf{M-CHMM ($\mathrm{M}=4$)} & $779\pm48$ & $599\pm28$ \\
\textbf{M-CHMM ($\mathrm{M}=5$)} & $772\pm46$ & $588\pm19$ \\
\textbf{M-CHMM ($\mathrm{M}=6$)} & $758\pm50$ & $618\pm34$ \\
\textbf{M-CHMM ($\mathrm{M}=7$)} & $765\pm55$ & $609\pm36$ \\
\textbf{M-CHMM ($\mathrm{M}=8$)} & $\mathbf{755\pm47}$ & $623\pm47$ \\
\textbf{M-CHMM ($\mathrm{M}=9$)} & $779\pm38$ & $626\pm36$ \\
\bottomrule
\end{tabular}
\label{table:comparison}
\end{table}

\subsection{Modeling the real-world data with the M-CHMM}

\begin{figure}[t]
    \centering
    \includegraphics[width=\columnwidth]{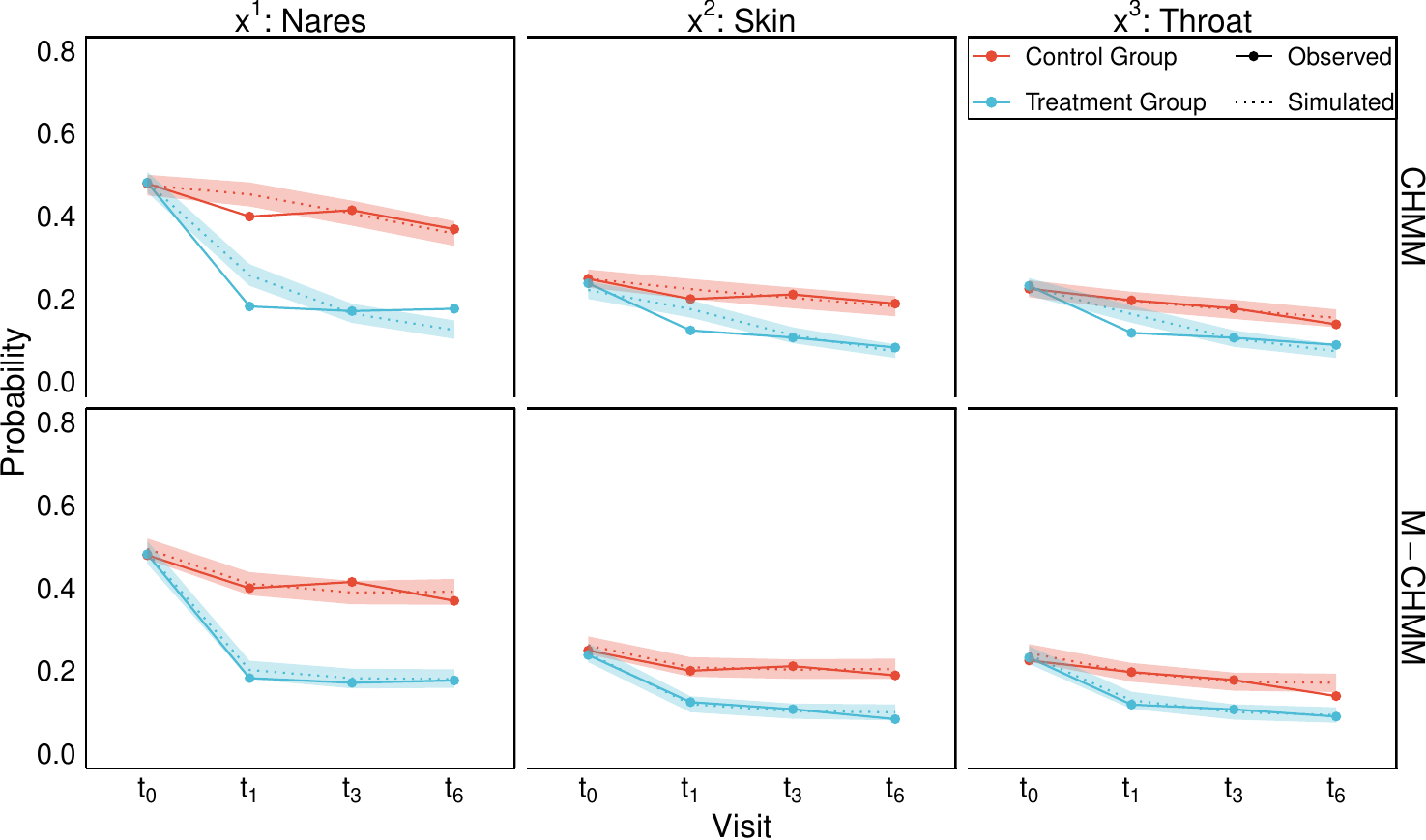}
    \caption{Posterior predictive checking of bacterial colonization for CHMM, and M-CHMM.}
    \label{fig:predictive}
\end{figure}

Here, we modeled the control and treatment groups separately with the M-CHMM to understand what kind of different patient subgroups they included. From this point onwards, we only used PF for inference since it was the most accurate and scalable in the previous experiments. \autoref{table:comparison} compares different numbers of clusters using negative log-likelihood (NLL) scores in the test set. The best predictive accuracy was obtained with 8 clusters in the control group and 3 clusters in the treatment group. \autoref{fig:predictive} shows the results of posterior predictive checking for CHMM and M-CHMM with the optimal number of clusters. We see that the CHMM fails to model the decrease in the probability of bacterial colonization in the different body parts over time, but the M-CHMM accurately captures it and successfully regenerates the observations. This demonstrates the flexibility of the M-CHMM to capture complex patterns in real-world time-series data by identifying groups of individuals with different dynamics.

\begin{figure}[b]
    \centering
    \includegraphics[width=\columnwidth]{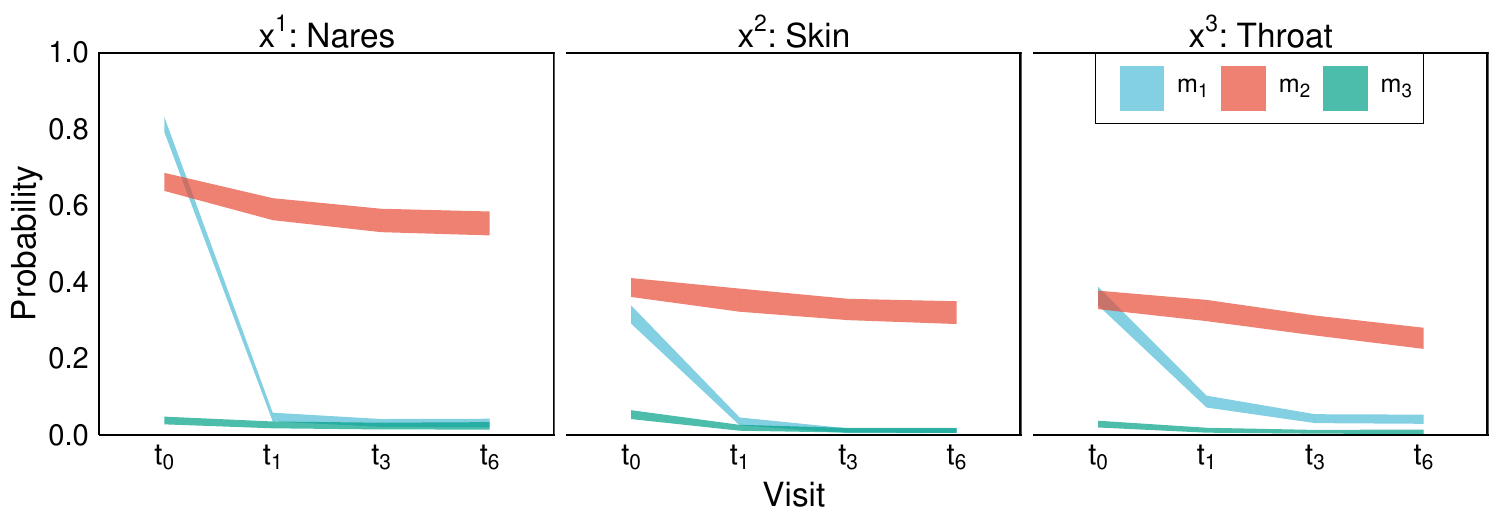}
    \caption{The dynamics of the clusters detected by M-CHMM with 3 clusters.}
    \label{fig:groups}
\end{figure}

\begin{figure}[t]
    \centering
    \includegraphics[width=\columnwidth]{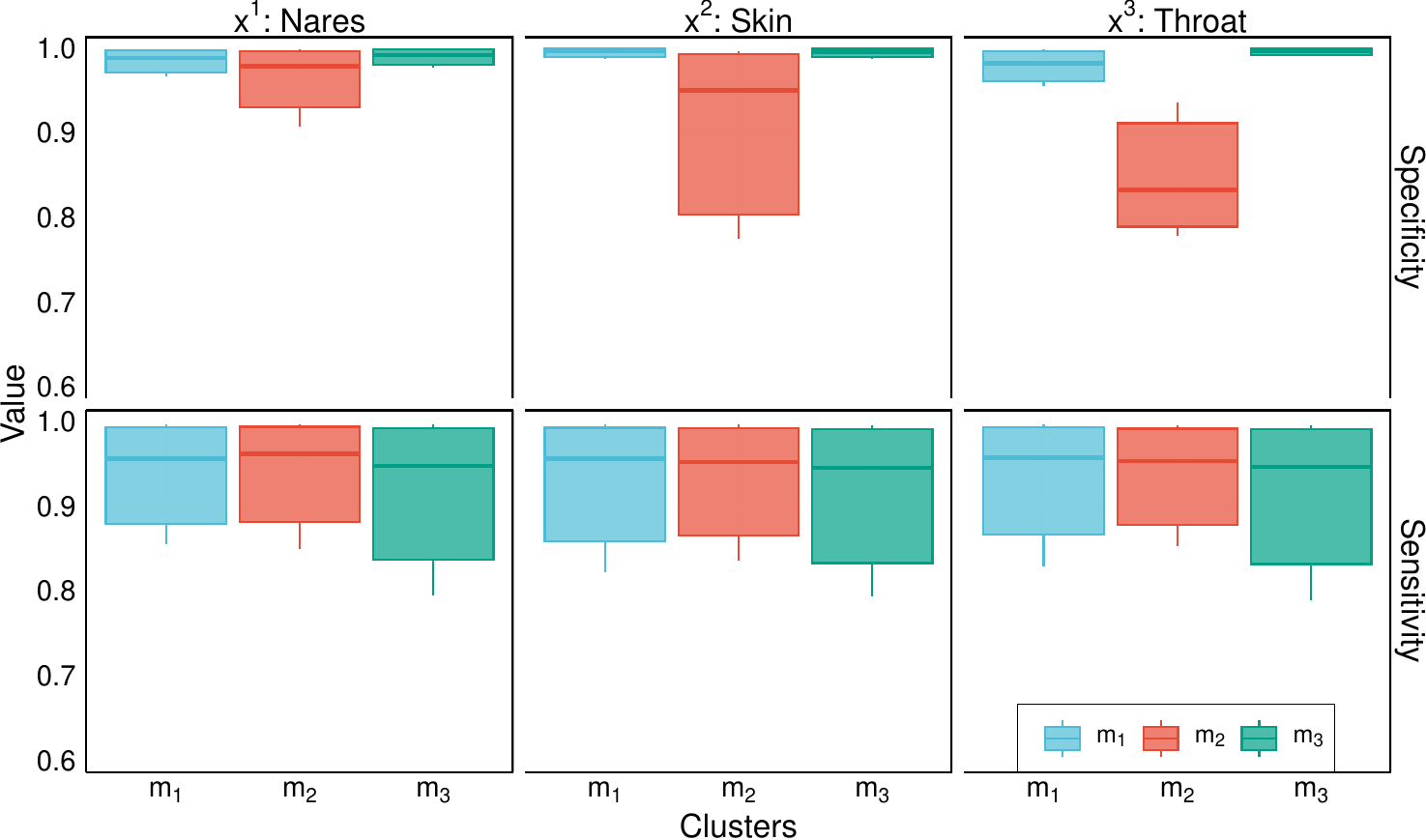}
    \caption{The specificity and sensitivity analysis of measurements for 3 clusters.}
    \label{fig:sensitivity}
\end{figure}

Next, we investigated the learned clusters within the treatment group (the control group is analyzed in \aref{section:additional}). \autoref{fig:groups} clearly shows one cluster of patients who responded well to the treatment (probability of bacterial colonization dropped rapidly) and another cluster where the bacterium was persistent (probability stayed constantly high), which reflects the medical understanding of \citet{huang2019decolonization}. The third cluster included patients colonized with a low probability already in the initial phase. \autoref{fig:sensitivity} show the sensitivity and specificity of the measurements. Overall, similar values compatible with literature are estimated for different clusters \citep{carr2018clinical}, with a slight shift of cluster $m_2$, for which a possible explanation is that the treatment may affect the measurement accuracy \citep{carr2018clinical}.

\section{Discussion}
\label{section:Discussion}

We proposed the M-CHMM as a flexible extension of the CHMM for analyses of nonexchangeable multivariate data from longitudinal clinical trials by identifying clusters of patients with different temporal dynamics. The CHMMs are ideal for systems consisting of multiple interacting latent processes where the aim is to learn about the dynamics of the system and interactions between its components. Many systems in epidemiology and healthcare match this description and, consequently, CHMMs have previously been found useful in multiple diverse biomedical applications: for disease co-morbidities \citep{maag2021modeling}, for hourly measurements of multiple vital signs of ICU patients \citep{pohle2021primer}, for accounting for genetic relatedness between individuals when detecting genetic copy-number variants jointly across individuals \citep{wang2019variational}, for disease transmission between individuals \citep{touloupou2020scalable} or between body sites of an individual \citep{poyraz2022modelling}, and for interactions between parasite species in a host \citep{sherlock2013coupled}. Our extension to mixtures of CHMMs is expected to be useful in all situations where the CHMMs are applicable, by providing increased flexibility. Furthermore, different systems, e.g., time-series from different patients, may not behave in a similar way, e.g., due to differences in treatment response. In such situations, the M-CHMM can identify differently behaving patients and thereby learn patterns that may provide insight into the data, as well as improve model fit and prediction accuracy, which can be beneficial in downstream use cases.

During our investigation, we found existing algorithms to sample the latent sequence of the CHMM either computationally demanding or not providing a way to estimate the likelihood, which prevented their robust application with M-CHMMs. To resolve this, we formulated two latent sequence samplers for CHMMs, and showed that they are more efficient and scalable than existing methods, and allow accurate likelihood estimation, facilitating inference with more complex models. By using them, we demonstrated the ability of the M-CHMM to model complex real-world multivariate time-series data with noisy and missing measurements. To facilitate a wide adoption, we provide the method as an R-package with C-code optimization of computationally heavy parts.\footnote{Available at \url{https://github.com/onurpoyraz/M-CHMM}.}

\acks{The authors gratefully acknowledge Susan S. Huang for providing the CLEAR (Changing Lives by Eradicating Antibiotic Resistance) dataset to be used in this study (available at \doi{10.6084/m9.figshare.19786678.v4}). This work was supported by the Academy of Finland (Flagship programme: Finnish Center for Artificial Intelligence FCAI, and grants 336033, 352986) and EU (H2020 grant 101016775 and NextGenerationEU).}

\bibliography{poyraz23}

\newpage
\appendix
\onecolumn

\section{Derivation of FFBS on CHMM}
\label{section:FFBS}

If we define $\alpha\left(\boldsymbol{\pi}_{t}^{\mathbb{C}}\right) = p\left(\boldsymbol{\pi}_{t}^{\mathbb{C}} \mid \mathbf{x}_{1 : t}^{\mathbb{C}}\right)$, the full {\it $\alpha$-recursion} or {\it forward recursion} of the single HMM in which all chains are handled jointly is derived as follows;
\begin{align}
\alpha\left(\boldsymbol{\pi}_{t}^{\mathbb{C}}\right) &= \sum_{\boldsymbol{\pi}_{t-1}^{\mathbb{C}}} p\left(\boldsymbol{\pi}_{t-1}^{\mathbb{C}}, \boldsymbol{\pi}_{t}^{\mathbb{C}} \mid \mathbf{x}_{1:t}^{\mathbb{C}}\right) \\
&\propto \sum_{\boldsymbol{\pi}_{t-1}^{\mathbb{C}}} p\left(\mathbf{x}_{t}^{\mathbb{C}} \mid \boldsymbol{\pi}_{t}^{\mathbb{C}}\right) p\left(\boldsymbol{\pi}_{t}^{\mathbb{C}} \mid \boldsymbol{\pi}_{t-1}^{\mathbb{C}}\right) p\left(\boldsymbol{\pi}_{t-1}^{\mathbb{C}} \mid \mathbf{x}_{1:t-1}^{\mathbb{C}}\right) \\
&= p\left(\mathbf{x}_{t}^{\mathbb{C}} \mid \boldsymbol{\pi}_{t}^{\mathbb{C}}\right) \sum_{\boldsymbol{\pi}_{t-1}^{\mathbb{C}}} p\left(\boldsymbol{\pi}_{t}^{\mathbb{C}} \mid \boldsymbol{\pi}_{t-1}^{\mathbb{C}}\right) p\left(\boldsymbol{\pi}_{t-1}^{\mathbb{C}} \mid \mathbf{x}_{1:t-1}^{\mathbb{C}}\right) \\
&= \underbrace{p\left(\mathbf{x}_{t}^{\mathbb{C}} \mid \boldsymbol{\pi}_{t}^{\mathbb{C}}\right)}_{\text {corrector}} \underbrace{\textstyle\sum_{\boldsymbol{\pi}_{t-1}^{\mathbb{C}}} p\left(\boldsymbol{\pi}_{t}^{\mathbb{C}} \mid \boldsymbol{\pi}_{t-1}^{\mathbb{C}}\right) \alpha\left(\boldsymbol{\pi}_{t-1}^{\mathbb{C}}\right)}_{\text{predictor}}. \label{eq:regular_alpha_recursion}
\end{align}

\section{Derivation of iFFBS}
\label{section:iFFBS}

If we define $\alpha\left(\pi_{t}^{c}\right) = p\left(\pi_{t}^{c} \mid \boldsymbol{\pi}_{1:t+1}^{\mathbf{\neg c}}, \mathbf{x}_{1: t}^{c}\right)$, the modified {\it $\alpha$-recursion} of CHMM using iFFBS algorithm can be derived for a single chain $c$ as following:
\begin{align} 
    \alpha\left(\pi_{t}^{c}\right) & = \sum_{\boldsymbol{\pi}_{t-1}^{c}} p\left(\pi_{t}^{c}, \pi_{t-1}^{c} \mid \mathbf{x}_{1 : t-1}^{c}, x_{t}^{c}, \boldsymbol{\pi}_{1:t}^{\mathbf{\neg c}}, \boldsymbol{\pi}_{t+1}^{\mathbf{\neg c}} \right) \\
    & \propto \sum_{\boldsymbol{\pi}_{t-1}^{c}} p\left(x_{t}^{c} \mid \pi_{t}^{c}\right) p\left(\pi_{t}^{c} \mid \boldsymbol{\pi}_{t-1}^{\mathbb{C}}\right) p\left(\pi_{t-1}^{c} \mid \boldsymbol{\pi}_{1:t}^{\mathbf{\neg c}}, \mathbf{x}_{1:t-1}^{c}\right) \prod_{\hat{c} \in \mathbb{C}\setminus c} p\left(\pi_{t+1}^{\hat{c}} \mid \boldsymbol{\pi}_{t}^{\mathbb{C}} \right) \\
    & = \underbrace{p\left(x_{t}^{c} \mid \pi_{t}^{c}\right)}_{\text{corrector}} \underbrace{\textstyle\sum_{\pi_{t-1}^{c}} p\left(\pi_{t}^{c} \mid \boldsymbol{\pi}_{t-1}^{\mathbb{C}}\right) \alpha\left(\pi_{t-1}^{c}\right)}_{\text{predictor}}  \underbrace{\textstyle\prod_{\hat{c}\in \mathbb{C}\setminus c} p\left(\pi_{t+1}^{\hat{c}} \mid \boldsymbol{\pi}_{t}^{\mathbb{C}} \right)}_{\text{modifying mass}} \label{eq:modified-filtering}
\end{align}
If we normalize $\alpha\left(\pi_{t}^{c}\right) = p\left(\pi_{t}^{c} \mid \boldsymbol{\pi}_{1:t+1}^{\mathbf{\neg c}}, \mathbf{x}_{1: t}^{c}\right)$ in two step in which, first before multiplying the {\it modifying mass} and after multiplication, the first normalization constant will be $\mathrm{Z}_t^c = p \left( x_t^c \mid \mathbf{x}_{1:t-1}^c, \boldsymbol{\pi}_{1:t}^{\neg c} \right)$. In iFFBS algorithm, marginal likelihood is not available but we heuristically approximate it as follows:
\begin{align}
    \mathcal{L}\left(\mathbf{x}_{1:\mathrm{T}}^{\mathbb{C}}\right)\overset{\Delta}{=}p\left(\mathbf{x}_{1:\mathrm{T}}^{\mathbb{C}}\right) \approx \prod_{t=1}^{\mathrm{T}} \prod_{c=1}^{\mathrm{C}} \mathrm{Z}_t^c.
\end{align}

\section{Design of the Transition Probabilities}
\label{section:transition}

We notice that the unnormalized transition log-probabilities $\mu_t^c$ are an additive function of the latent states of the other chains, and transition probabilities $\tau_t^c$ are obtained by a normalization of the exponential function. Hence we can write:
\begin{align}
p(\pi_t^c\mid \boldsymbol{\pi}_{t-1}^{\mathbb{C}})&\propto e^{\beta_0^{c\leftarrow{c}} + \sum_{\hat{c} \in \mathbb{C} \setminus c} \sum_{k \in \mathbb{K}} \beta_k^{c\leftarrow{\hat{c}}} \mathbb{I} \left[ \pi_{t-1}^{\hat{c}} = k \right]}\\
&= e^{\beta_0^{c\leftarrow{c}}} \prod_{\hat{c} \in \mathbb{C} \setminus c}\prod_{k \in \mathbb{K}} e^{\beta_k^{c\leftarrow{\hat{c}}} \mathbb{I} \left[ \pi_{t-1}^{\hat{c}} = k \right]}\\
&\eqdef f_0(\pi_t^c,\pi_{t-1}^{c}) \prod_{\hat{c} \in \mathbb{C} \setminus c} f_{\pi_{t-1}^{\hat{c}}}(\pi_t^c, \pi_{t-1}^{c}).\label{eq:transition_factorization}
\end{align}
\autoref{eq:transition_factorization} means that the transition probability from $\pi_{t-1}^{c}$ to $\pi_{t}^{c}$ in chain $c$ factorizes into (an element-wise) product of matrices over chains, where the matrix corresponding to each chain $\hat{c}\not=c$ depends on the state $\pi_{t-1}^{\hat{c}}$ of the chain $\hat{c}$ in the previous time-step.

\section{Derivation of the Factorized FFBS (fFFBS)}
\label{section:fFFBS}

\subsection{Forward Filtering}
If we define $\alpha\left(\pi_{t}^c\right) = p\left(\pi_t^{c} \mid x_t^c, \mathbf{x}_{1:t-1}^{\mathbb{C}}\right)$ given factorization assumption defined in \autoref{eq:factorization}, the approximated {\it $\alpha$-recursion} for each chain can be derived as follows;
\begin{align}
\alpha(\pi_{t}^c) =& \sum_{\boldsymbol{\pi}_{t-1}^{\mathbb{C}}} p(\pi_{t}^{c}, \boldsymbol{\pi}_{t-1}^{\mathbb{C}} \mid x_t^c, \mathbf{x}_{1:t-1}^{\mathbb{C}}) \\
\propto& \sum_{\boldsymbol{\pi}_{t-1}^{\mathbb{C}}} p\left(x_{t}^{c} \mid \pi_{t}^{c}\right) p\left(\pi_{t}^{c} \mid \boldsymbol{\pi}_{t-1}^{\mathbb{C}}\right) p(\boldsymbol{\pi}_{t-1}^{\mathbb{C}} \mid \mathbf{x}_{1:t-1}^{\mathbb{C}}) \\
=& p\left(x_{t}^{c} \mid \pi_{t}^{c}\right) \sum_{\boldsymbol{\pi}_{t-1}^{\mathbb{C}}} p\left(\pi_{t}^{c} \mid \boldsymbol{\pi}_{t-1}^{\mathbb{C}}\right) p\left(\boldsymbol{\pi}_{t-1}^{\mathbb{C}} \mid \mathbf{x}_{1:t-1}^{\mathbb{C}}\right) \\
\approx& p\left(x_{t}^{c} \mid \pi_{t}^{c}\right) \sum_{\boldsymbol{\pi}_{t-1}^{\mathbb{C}}} p\left(\pi_{t}^{c} \mid \boldsymbol{\pi}_{t-1}^{\mathbb{C}}\right) \prod_{\hat{c} \in \mathbb{C}} \underbrace{p\left(\pi_{t-1}^{\hat{c}} \mid x_{t-1}^{\hat{c}}, \mathbf{x}_{1:t-2}^{\mathbb{C}}\right)}_{\alpha(\pi_{t-1}^{\hat{c}})} \\
=& \underbrace{p\left(x_{t}^{c} \mid \pi_{t}^{c}\right)}_{\epsilon^c} \sum_{\pi_{t-1}^{c}} \alpha(\pi_{t-1}^c) \underbrace{\sum_{\boldsymbol{\pi}_{t-1}^{\mathbb{C}\setminus c}} p\left(\pi_{t}^{c} \mid \boldsymbol{\pi}_{t-1}^{\mathbb{C}}\right) \prod_{\hat{c} \in \mathbb{C }\setminus c} \alpha(\pi_{t-1}^{\hat{c}})}_{\tau_t^c} \label{eq:chain-forward-supplement}\\
=& \epsilon^c \sum_{\pi_{t-1}^{c}}\alpha(\pi_{t-1}^c) \tau_t^c.
\label{eq:factorized_alpha_recursion}
\end{align}
Here, we see that $\tau_t^c$ has a role analogous to the transition matrix in the regular forward-filtering algorithm, i.e., the probability $p\left(\boldsymbol{\pi}_{t}^{\mathbb{C}} \mid \boldsymbol{\pi}_{t-1}^{\mathbb{C}}\right)$ in \autoref{eq:regular_alpha_recursion}. In detail, it is a matrix specifying the probabilities of moving from $\pi_{t-1}^{c}$ to $\pi_{t}^{c}$, weighing the transitions by the probabilities of the states $\pi_{t-1}^{\hat{c}}$ in the other chains $\hat{c}$.

\subsection{Reinterpretation of Transition Matrix}
By using the proportionality in \autoref{eq:reinterpretation} we can define dynamic transition matrices $\tau_t^c$ in \autoref{eq:chain-forward} as follows:
\begin{align}
    \tau_t^c &= \sum_{\boldsymbol{\pi}_{t-1}^{\mathbb{C}\setminus c}} p\left(\pi_{t}^{c} \mid \boldsymbol{\pi}_{t-1}^{\mathbb{C}}\right) \prod_{\hat{c} \in \mathbb{C }\setminus c} \alpha(\pi_{t-1}^{\hat{c}}) \\
    &\propto \sum_{\boldsymbol{\pi}_{t-1}^{\mathbb{C}\setminus c}} f_0(\pi_t^c,\pi_{t-1}^{c}) \prod_{\hat{c} \in \mathbb{C} \setminus c} f_{\pi_{t-1}^{\hat{c}}}(\pi_t^c, \pi_{t-1}^{c}) \prod_{\hat{c} \in \mathbb{C }\setminus c} \alpha(\pi_{t-1}^{\hat{c}}) \\
    &= f_0(\pi_t^c,\pi_{t-1}^{c}) \sum_{\boldsymbol{\pi}_{t-1}^{\mathbb{C}\setminus c}}\prod_{\hat{c} \in \mathbb{C} \setminus c} f_{\pi_{t-1}^{\hat{c}}}(\pi_t^c, \pi_{t-1}^{c}) \alpha(\pi_{t-1}^{\hat{c}}) \\
    &= f_0(\pi_t^c,\pi_{t-1}^{c}) \prod_{\hat{c} \in \mathbb{C} \setminus c}  \sum_{\boldsymbol{\pi}_{t-1}^{\hat{c}}} f_{\pi_{t-1}^{\hat{c}}}(\pi_t^c, \pi_{t-1}^{c}) \alpha(\pi_{t-1}^{\hat{c}})
\end{align}
where rows of $\tau_t^c$ must sum to one by definition. So, we can normalize it at the final step, and $\tau_t^c$ will have an equality, not proportionality. 

\subsection{Time Reversed Transition Probability}
Given \autoref{eq:factorization} at each $t$ and \autoref{eq:reinterpretation}, we can expand the time reversed transition probabilities as follows:
\begin{align}
    p(\boldsymbol{\pi}_t^{\mathbb{C}} \mid \boldsymbol{\pi}_{t+1}^{\mathbb{C}}, \mathbf{x}_{1:t}^{\mathbb{C}}) &\propto p(\boldsymbol{\pi}_{t+1}^{\mathbb{C}} \mid \boldsymbol{\pi}_{t}^{\mathbb{C}}) p(\boldsymbol{\pi}_t^{\mathbb{C}} \mid \mathbf{x}_{1:t}^{\mathbb{C}}) \\
    &= p(\boldsymbol{\pi}_{t+1}^{\mathbb{C}} \mid \boldsymbol{\pi}_{t}^{\mathbb{C}}) \alpha\left(\boldsymbol{\pi}_{t}^{\mathbb{C}}\right) \\
    &\approx \prod_{c\in \mathbb{C}} p(\pi_{t+1}^{c} \mid \boldsymbol{\pi}_{t}^{\mathbb{C}}) \alpha\left(\pi_{t}^{c}\right) \\
    &\propto \prod_{c\in \mathbb{C}} \alpha\left(\pi_{t}^{c}\right) \left[\textstyle f_0(\pi_{t+1}^c,\pi_{t}^{c}) \prod_{\hat{c} \in \mathbb{C} \setminus c} f_{\pi_{t}^{\hat{c}}}(\pi_{t+1}^c, \pi_{t}^{c})\right] \\
    &= \prod_{c\in \mathbb{C}} \alpha\left(\pi_{t}^{c}\right) \left[\textstyle f_0(\pi_{t+1}^c,\pi_{t}^{c}) \prod_{\hat{c} \in \mathbb{C} \setminus c} f_{\pi_{t}^{c}}(\pi_{t+1}^{\hat{c}}, \pi_{t}^{\hat{c}})\right]
\end{align}
Here, in final step, we rearranged the factors, where the factors are grouped by the outgoing factors. We see that the factorization assumption leads to efficient computation by allowing us to operate on matrices whose size equals the size of a transition matrix for one chain only.

\section{Adaptive Metropolis-Hastings within Gibbs Sampling}
\label{section:MHwithinGibbs}

We have described the design specifications transition matrix of the CHMM in Section~3.2, in which $\boldsymbol{\beta}^{c}$ is defined as a set of matrices. Here, by abusing the notation, we consider all the parameters related to a chain $c$ as a single vector, denoted as $\boldsymbol{\beta}^{c}$. Since transitions are changing at each time step, it is not feasible to calculate their sufficient statistics as in HMM. To resolve this, we sample the $\boldsymbol{\beta}^{c*}$ parameters using a Metropolis-Hastings(MH) step within the Gibbs sampler conditional on the latent states of all the chains. The transition parameters for each chain $\boldsymbol{\tau}^c$ are fully determined by the $\boldsymbol{\beta}^{c}$ parameters. To get a draw from $p\left(\boldsymbol{\beta}^{c} \mid \boldsymbol{\pi}_{1:\mathrm{T}}^{c}, \boldsymbol{\pi}_{1:\mathrm{T}}^{\mathbf{\neg c}}\right)$ we use the following steps;
\begin{enumerate}
    \item Make a proposal $\boldsymbol{\beta}^{c*} \sim q\left(\boldsymbol{\beta}^{c}\right)$
    \item Transform $\boldsymbol{\beta}^{c*}$ samples to transition probabilities $\boldsymbol{\tau}_{1:\mathrm{T}}^{c*}$ according to latent states of the other chains, $\boldsymbol{\pi}^{\mathbf{\neg c}}$
    \item Accept the proposal $\boldsymbol{\beta}^{c*}$ with probability
    \begin{align}
        \alpha \left(\boldsymbol{\tau}_{1:\mathrm{T}}^{c*}\right) = \min \left \{ 1, \frac{p\left(\boldsymbol{\beta}^{c*} \mid \boldsymbol{\pi}_{1:\mathrm{T}}^{c}, \boldsymbol{\pi}_{1:\mathrm{T}}^{\mathbf{\neg c}}\right) q\left(\boldsymbol{\beta}^{c} \mid \boldsymbol{\beta}^{c*}\right)} {p\left(\boldsymbol{\beta}^{c} \mid \boldsymbol{\pi}_{1:\mathrm{T}}^{c}, \boldsymbol{\pi}_{1:\mathrm{T}}^{\mathbf{\neg c}}\right)q\left(\boldsymbol{\beta}^{c*} \mid \boldsymbol{\beta}^{c} \right)} \right \} \label{eq:acceptance}
    \end{align}
    where
    \begin{align}
        &p\left(\boldsymbol{\beta^{c}} \mid \boldsymbol{\pi}_{1:\mathrm{T}}^{c}, \boldsymbol{\pi}_{1:\mathrm{T}}^{\mathbf{\neg c}}\right) \propto p\left(\boldsymbol{\pi}_{1:\mathrm{T}}^{c} \mid \boldsymbol{\beta}^{c}, \boldsymbol{\pi}_{1:\mathrm{T}}^{\mathbf{\neg c}}\right)p\left(\boldsymbol{\beta}^{c}\right)
        \label{eq:post}
    \end{align}
\end{enumerate}
The first quantity on the right-hand side of the \autoref{eq:post} can be directly calculated with the transition parameters $\boldsymbol{\tau}_{1:\tau}^{c}$ for the corresponding chain. We used the Gaussian proposal in which the previously sampled $\boldsymbol{\beta}^{c}$ is the mean such that $q\left(\boldsymbol{\beta}^{c*} \mid \boldsymbol{\beta}^{c}\right) = \mathcal{N}\left(\boldsymbol{\beta}^{c*} \mid \boldsymbol{\beta}^{c}, \psi^2 \Sigma \right)$ where $\psi$ is scaling factor and $\Sigma$ is covariance matrix. We initially set $\psi \approx 2.38/\sqrt{d}$ as Metropolis jumping scaling factor since it is theoretically the most efficient scaling factor \citep{gelman2013bayesian}, where $d$ is the dimension of the sampling. As a variance parameter, we used a fixed diagonal covariance matrix $\Sigma = 0.01\times I$, where $I$ is the identity matrix, (which corresponds to a step size giving the optimal acceptance rate of $\approx 23\%$ \citep{roberts2001optimal}) during the warm-up period and then used the online covariance matrix estimated from the previous samples \cite{dasgupta2007line}. Then we adaptively scaled the scaling factor $\psi$ as described in \citet{sherlock2010random}. Since the proposal distribution is symmetric (i.e., Normal), $q\left(\boldsymbol{\beta}^{c} \mid \boldsymbol{\beta}^{c*} \right)$ and $q\left(\boldsymbol{\beta}^{c*} \mid \boldsymbol{\beta}^{c} \right)$ in \autoref{eq:acceptance}, cancel out each other.

\section{Implementation details}
\label{section:implementation}

We set the prior of $\beta_0^c$ as $N(\beta_0^c \mid 0, 1)$ for any chain, which is almost uninformative so that the estimates are not affected strongly by the prior. For the rest of the $\beta$ parameters denoted by $\beta_k$, we used sparsity encouraging Horseshoe prior with mean and scale parameters of $0$ and $0.25$, respectively. We used a uniform prior on the initial state probabilities $\pi_0$ and weak Dirichlet priors for the rows of emission probabilities $E$ such that we set the value to $30$ for specificity and $15$ for sensitivity. We set the rest of the emission priors to $1$, which corresponds to the uniform prior. In such a formulation, except for the initialization, the prior has a negligible effect because it is summed with observation counts during inference. We drew 20,000 MCMC samples and set the warm-up length as 10,000. Posterior probabilities are calculated using the rest of MCMC samples.

\section{Additional Results}
\label{section:additional}

In this section, we provided additional analysis and results which supports the outcomes of the manuscript.

\begin{Supplementary Figure}[!p]
    \centering
    \includegraphics[width=\linewidth]{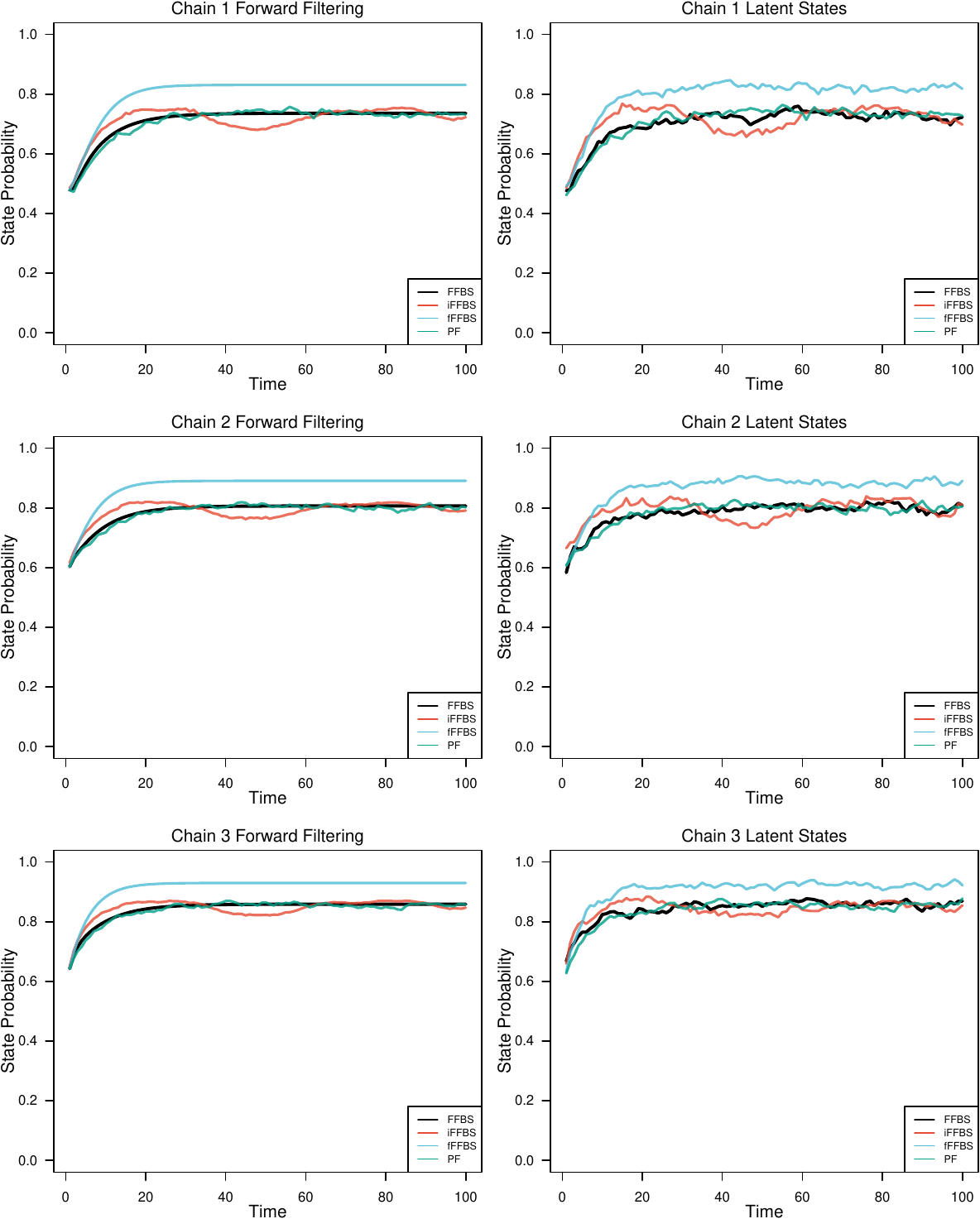}
    \caption{
    Error accumulation without observation.
    }
    \label{fig:EA-NULL}
\end{Supplementary Figure}

\begin{Supplementary Figure}[!p]
    \centering
    \includegraphics[width=\linewidth]{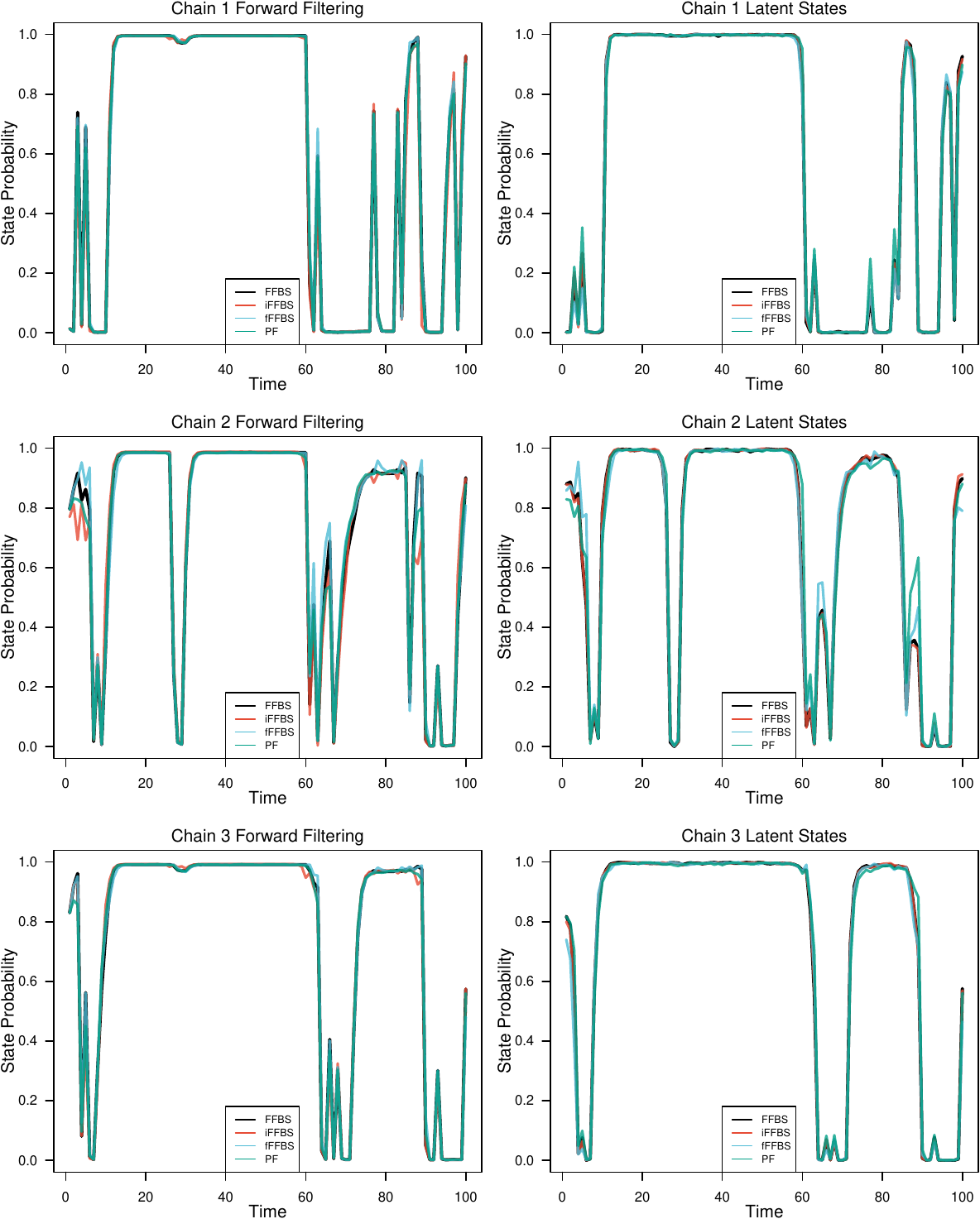}
    \caption{
    Error accumulation with observation.
    }
    \label{fig:EA}
\end{Supplementary Figure}

\begin{Supplementary Figure}[!p]
    \centering
    \includegraphics[width=0.98\linewidth]{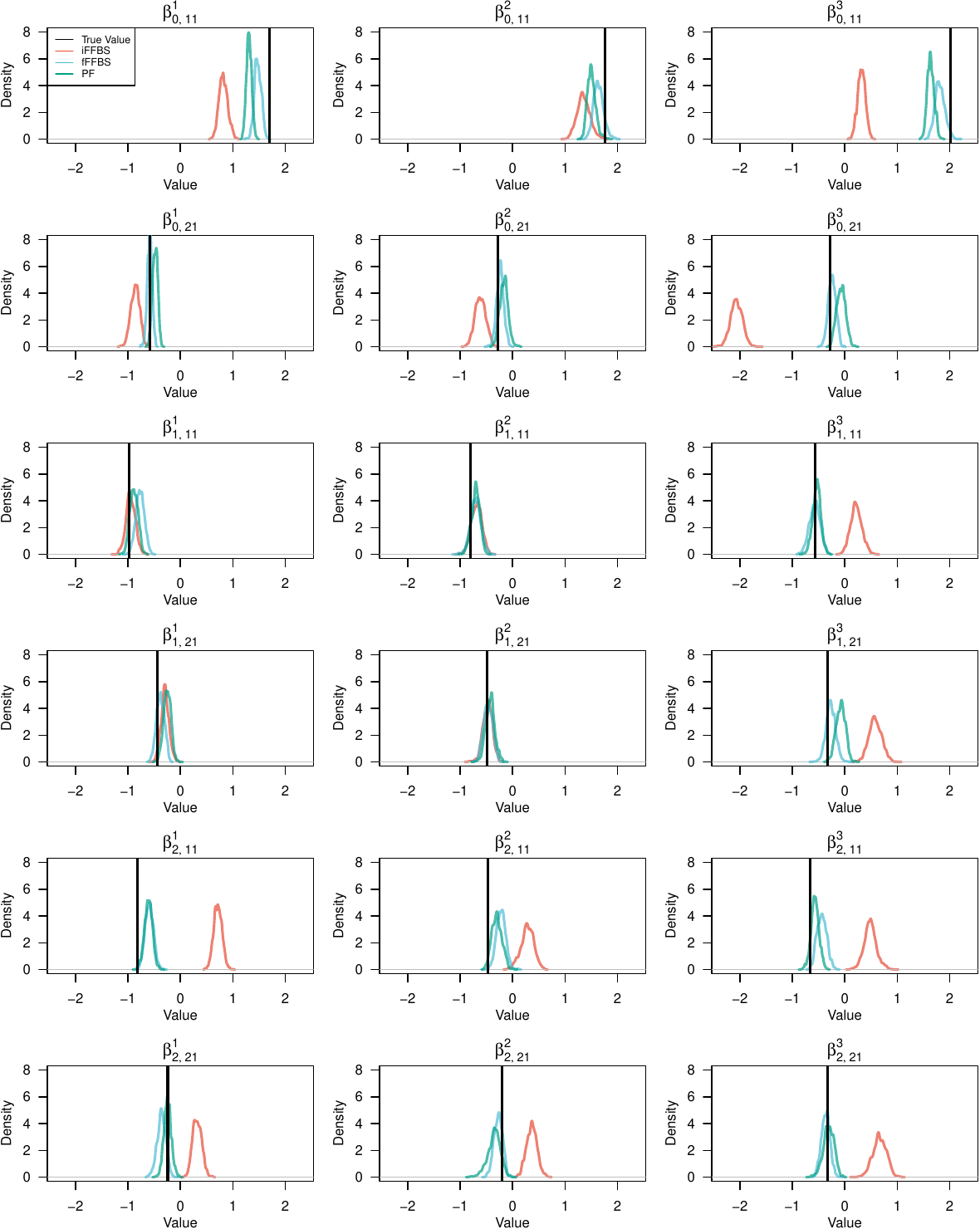}
    \caption{
    Convergence of $\boldsymbol{\beta}$ parameters for each chain with different latent sequence samplers.
    }
    \label{fig:mcmc-posterior-appendix}
\end{Supplementary Figure}

\begin{Supplementary Figure}[!p]
    \centering
    \includegraphics[width=\textwidth]{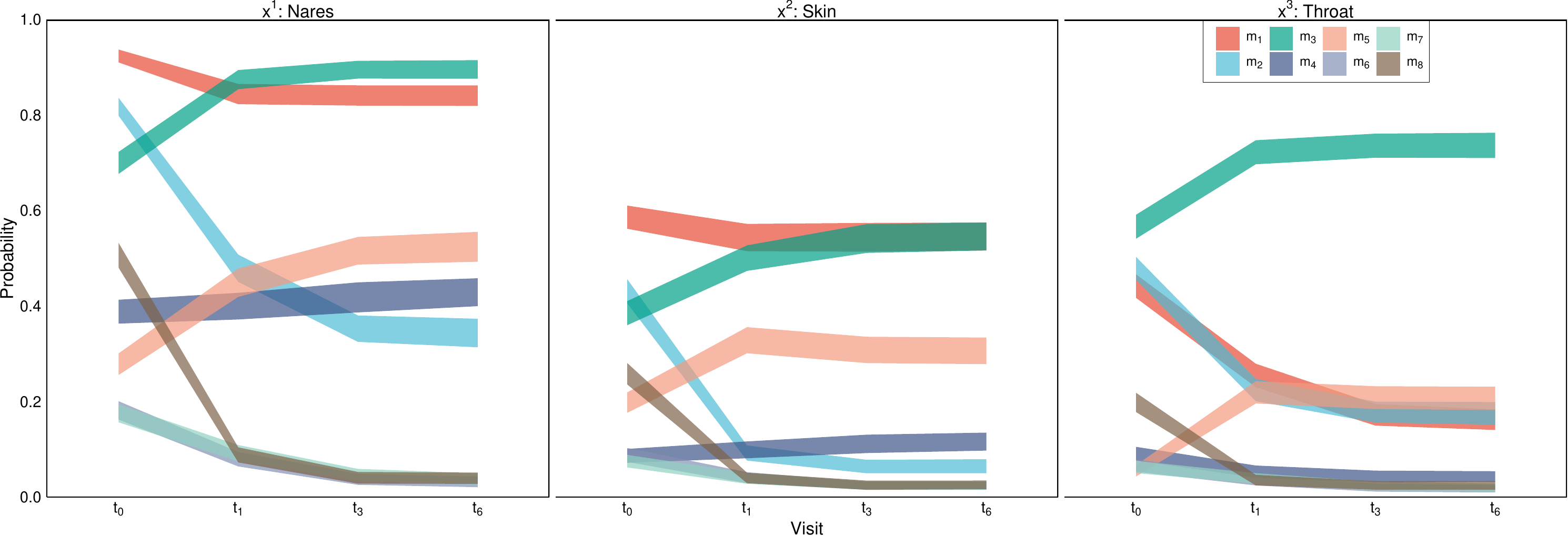}
    \caption{The dynamics of the clusters in the control group detected by M-CHMM with 8 clusters where $n=130,117,121,144,135,154,122$ and $129$, respectively, for $m_1, m_2, m_3, m_4, m_5, m_6, m_7$ and $m_8$. The color scheme is the same with \autoref{fig:sensitivity-edu}.}
    \label{fig:groups-edu}
\end{Supplementary Figure}

\begin{Supplementary Figure}[!p]
    \centering
    \includegraphics[width=\textwidth]{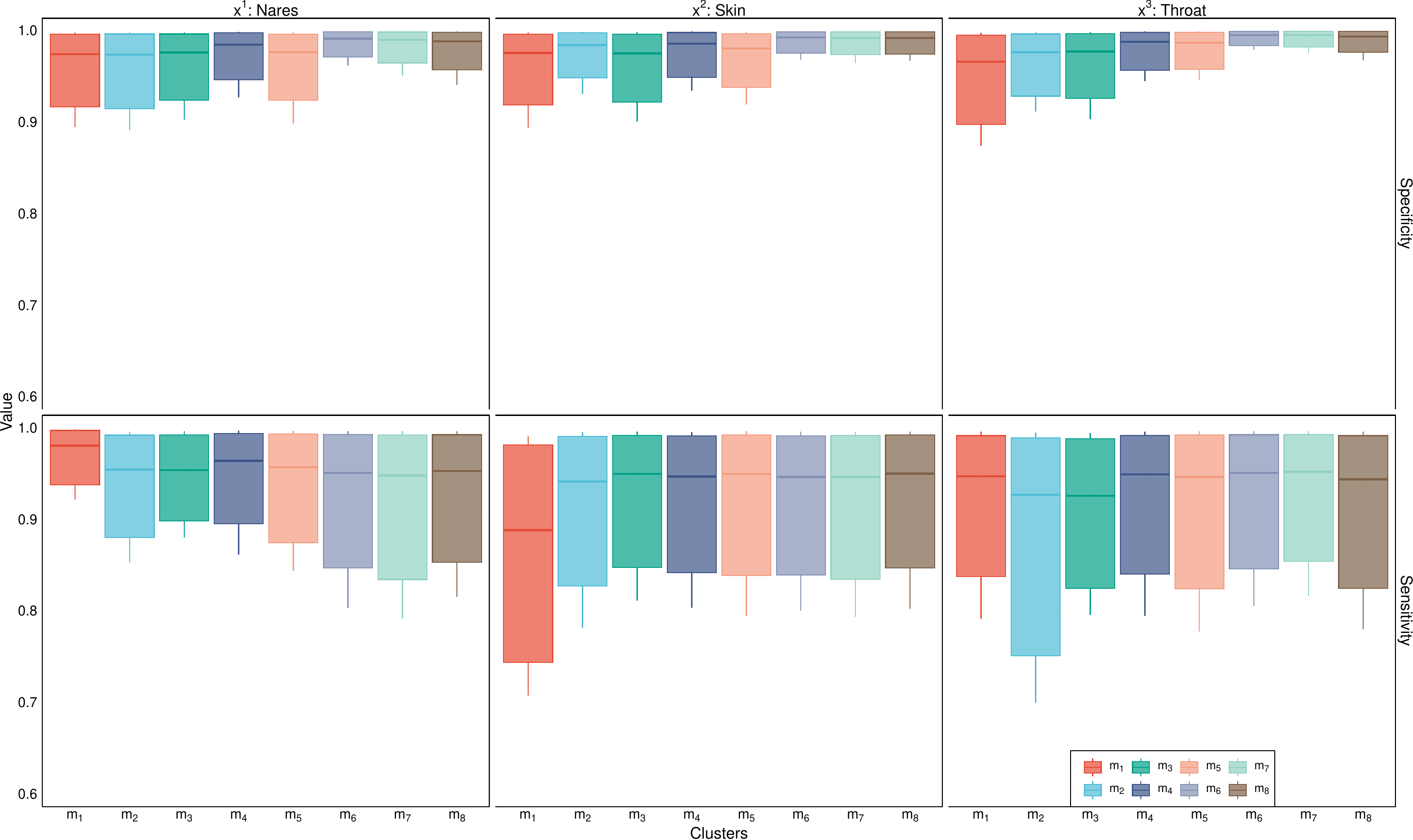}
    \caption{The specificity and sensitivity analysis of measurements in control group by cluster in M-CHMM, where $n=130,117,121,144,135,154,122$ and $129$, respectively, for $m_1, m_2, m_3, m_4, m_5, m_6, m_7$ and $m_8$. The color scheme is the same with \autoref{fig:groups-edu}.}
    \label{fig:sensitivity-edu}
\end{Supplementary Figure}

\begin{Supplementary Table}[ht]
\centering
\caption{Effective sample size (ESS) of each model with 10000 MCMC samples.}
\begin{tabular}{l c c c c c c c c}
\toprule
 & \multicolumn{4}{c}{\textbf{Control Group}} & \multicolumn{4}{c}{\textbf{Treatment Group}}\\
\cmidrule(lr){1-1} \cmidrule(lr){2-5} \cmidrule(lr){6-9}
\textbf{Model} & \textbf{FFBS} & \textbf{iFFBS} & \textbf{fFFBS} & \textbf{PF} & \textbf{FFBS} & \textbf{iFFBS} & \textbf{fFFBS} & \textbf{PF}\\
\midrule
\textbf{CHMM} & $1101$ & $2253$ & $708$ & $3147$ & $2115$ & $3099$ & $1482$ & $1756$ \\
\textbf{M-CHMM ($M=2$)} & $966$ & $664$ & $676$ & $958$ & $529$ & $576$ & $787$ & $915$ \\
\textbf{M-CHMM ($M=3$)} & $499$ & $249$ & $503$ & $627$ & $610$ & $549$ & $695$ & $786$ \\
\textbf{M-CHMM ($M=4$)} & $550$ & $203$ & $242$ & $343$ & $494$ & $409$ & $378$ & $456$ \\
\textbf{M-CHMM ($M=5$)} & $366$ & $235$ & $483$ & $374$ & $518$ & $264$ & $263$ & $387$ \\
\textbf{M-CHMM ($M=6$)} & $415$ & $247$ & $366$ & $341$ & $461$ & $307$ & $320$ & $355$ \\
\textbf{M-CHMM ($M=7$)} & $467$ & $247$ & $339$ & $300$ & $509$ & $327$ & $412$ & $421$ \\
\textbf{M-CHMM ($M=8$)} & $439$ & $264$ & $350$ & $335$ & $382$ & $307$ & $321$ & $360$ \\
\textbf{M-CHMM ($M=9$)} & $463$ & $283$ & $317$ & $360$ & $482$ & $343$ & $367$ & $435$ \\
\bottomrule
\end{tabular}
\label{table:ess}
\end{Supplementary Table}

\end{document}